\documentclass[smallcondensed]{svjour3}
\usepackage{graphicx}

\usepackage{times}
\usepackage{amsmath,bm}
\usepackage{xfrac}
\usepackage{xspace}
\usepackage{paralist}
\usepackage{algorithmic}
\usepackage{amsfonts}
\usepackage[inoutnumbered,linesnumbered,ruled,vlined]{algorithm2e}
\usepackage{color}
\usepackage{array}
\usepackage{multirow}
\usepackage{multicol}
\usepackage{rotating}
\usepackage{tablefootnote}
\usepackage{enumerate}
\usepackage{hyperref}
\usepackage{subcaption}
\usepackage{footmisc}

\usepackage{xcolor}
\usepackage[misc,geometry]{ifsym} 

\newcommand{\tabincell}[2][c]{\begin{tabular}[#1]{@{}c@{}}#2\end{tabular}}
\newcommand{\ejr}[1]{{\color{black}#1}}

\newcommand{\sroiq}{\ensuremath{\mathcal{SROIQ}}}
\newcommand{\mr}[1]{{\color{black}#1}}

\newcommand{\owlvec}{\textsf{OWL2Vec}$^{*}$\xspace}
\newcommand{\owl}{OWL\xspace}
\newcommand{\ie}{\textit{i}.\textit{e}.,\xspace}
\newcommand{\eg}{\textit{e}.\textit{g}.,\xspace}
\newcommand{\nsp}{\negthickspace}

\begin{document}
%
\title{\owlvec: Embedding of OWL Ontologies}

%
%

\author{Jiaoyan Chen         \and
        Pan Hu \and
        Ernesto Jimenez-Ruiz \and
        Ole Magnus Holter \and 
        Denvar Antonyrajah \and 
        Ian Horrocks
}


\institute{Jiaoyan Chen, Pan Hu and Ian Horrocks \at Department of Computer Science, University of Oxford, UK
           \and
           Ernesto Jimenez-Ruiz \at City, University of London, UK and
              Department of Informatics,  University of Oslo, Norway
           \and
           Ole Magnus Holter \at
              Department of Informatics,  University of Oslo, Norway
           \and
           Denvar Antonyrajah \at Samsung Research, UK 
}

\date{Received: date / Accepted: date}

\maketitle              
\begin{abstract}
Semantic embedding of knowledge graphs has been widely studied and used for prediction and statistical analysis tasks across various domains such as Natural Language Processing and the Semantic Web.
However, less attention has been paid to developing robust methods for embedding OWL (Web Ontology Language) ontologies \mr{which can express a much wider range of semantics than knowledge graphs and have been widely adopted in domains such as bioinformatics. 
In this paper, we propose a random walk and word embedding based} ontology embedding method named \owlvec, which encodes the semantics of an OWL ontology by taking into account its graph structure, lexical information and logical constructors. 
Our empirical evaluation with three real world datasets suggests that \owlvec benefits from these three different aspects of an ontology in class membership prediction and class subsumption prediction tasks. Furthermore, \owlvec often significantly outperforms the state-of-the-art methods in our experiments.

\keywords{Ontology \and Ontology Embedding \and Word Embedding \and Web Ontology Language \and \owlvec \and Ontology Completion}
\end{abstract}
\setcounter{footnote}{0} 
\section{Introduction}

In recent years, the semantic embedding of knowledge graphs (KGs)
has been widely investigated \cite{wang2017knowledge}.
The objective of such embeddings is to represent in a vector space KG components 
such as entities and relations in a way that captures the structure of the graph. 
Various kinds of KG embedding algorithms have been proposed and successfully applied to  
KG refinement (e.g., link prediction \cite{rossi2020knowledge} and entity alignment \cite{sun2020benchmarking}), recommendation systems \cite{ristoski2019rdf2vec}, zero-shot learning \cite{chen2020ontology,wang2018zero}, interaction prediction in bioinformatics \cite{smaili2018onto2vec,myklebust2019knowledge}, and so on.
However, most of these algorithms focus on creating embeddings for multi-relational graphs composed of triples in RDF (Resource Description Framework)\footnote{\url{https://www.w3.org/RDF/}} form such as $\langle$\textit{England}, \textit{isPartOf}, \textit{UK}$\rangle$ and $\langle$\textit{UK}, \textit{hasCapital}, \textit{London}$\rangle$.
They do not deal with OWL\footnote{\url{https://www.w3.org/OWL/}} ontologies (or ontological schemas in OWL) which include not only graph structures\footnote{In this paper an ontology's graph structure includes the relation between instances (e.g., \textit{isPartOf}) as in RDF KGs, the subsumption relation between classes (i.e., the class hierarchy defined by \textit{rdfs:subClassOf}), the membership relation between instances and classes, and the relation between instances and literals.},
but also logic constructors such as class disjointness, existential and universal quantification (e.g., \textit{a country must have at least one city as its capital}), 
and meta data such as the synonyms, definitions and comments of a class.
OWL ontologies have been widely used in many domains such as bioinformatics, the Semantic Web \mr{and Linked Data} \cite{myklebust2019knowledge,horrocks2008ontologies}.
They are capable of expressing complex domain knowledge and managing large scale domain vocabularies,
and can often improve the quality and usability of the KG \cite{paulheim2015serving,chen2020correcting}.

Inspired by the success of KG embeddings, more recently there has been a growing interest in  embedding simple ontological schemas consisting, e.g., of hierarchical classes, and property domain and range \cite{hao2019universal,moon2017learning,alshargi2018metrics,guan2019knowledge};
however, these methods rely on having a large number of facts (i.e., an ABox), 
and do not support more expressive OWL ontologies which contain some widely used logic constructors such as the class disjointness and the existential quantification mentioned above.
Embeddings for OWL ontologies have started to receive some attention recently.
Kulmanov et al. \cite{kulmanov2019embeddings} and  Garg et al.~\cite{garg2019quantum} proposed to model the semantics of the logic constructor by geometric learning, but their models only support some of the logic constructors from the description logics (DLs) $\mathcal{EL}^{++}$ (which is closely related to \owl EL -- a fragment of \owl) and $\mathcal{ALC}$,  respectively.
Moreover, both methods consider only the logical and graph structure of an ontology, and ignore its lexical information that widely exists in the meta data (e.g., \textit{rdfs:label} and \textit{rdfs:comment} triples).
OPA2Vec \cite{smaili2018opa2vec} considers the ontology's lexical information by learning a \mr{word embedding} model which encodes statistical correlations between items in a corpus.
However, it treats each axiom as a sentence and fails to explore and utilize the semantic relationships between axioms.
OWL2Vec \cite{holter2019embedding}, which is our very preliminary work before \owlvec, captures the semantics of \owl ontologies by exploring the neighborhoods of classes.
This was shown to be quite effective, but it does not fully exploit \mr{the graph structure, the lexical semantics, or the (onto)logical semantics} available in OWL ontologies.
In this work we have extended OWL2Vec in order to provide a more general and robust \owl ontology embedding framework which we call \owlvec. \owlvec exploits an \owl (or OWL 2) ontology by walking over its graph forms 
and generates a corpus of three documents that capture different aspects of the semantics of the ontology: \textit{(i)} the graph structure and the logic constructors, \textit{(ii)} the lexical information (\eg entity names, comments and definitions), and
\textit{(iii)} a combination of the lexical information, graph structure and logical constructors.
Finally, \owlvec uses a \mr{word embedding} model to create embeddings of both entities and words from the generated corpus. 
Note that the \owlvec framework is compatible \mr{with different word embedding methods and their different settings, although the current implementation adopts \textit{Word2Vec} \cite{mikolov2013distributed} and its skip-gram architecture.}

We have evaluated \owlvec in two case studies -- class membership prediction and class subsumption prediction, using three large scale real world ontologies -- a healthy lifestyle ontology named HeLis \cite{dragoni2018helis}, a food ontology named FoodOn \cite{dooley2018foodon} and the Gene Ontology (GO) \cite{gene2008gene}. 
In the case studies we empirically analyze the impact of \textit{(i)} different document and embedding settings which correspond to combinations of the semantics of the graph structure, lexical information and logic constructors, \textit{(ii)} different graph structure exploration settings (e.g., the transformation methods from an OWL ontology to an RDF graph, and the graph walking strategies), \textit{(iii)} ontology entailment reasoning, and \textit{(iv)} \mr{word embedding} pre-training.
The results suggest that \owlvec can achieve significantly better performance than the baselines including the state-of-the-art ontology embeddings \cite{kulmanov2019embeddings,garg2019quantum,smaili2018opa2vec,holter2019embedding}, some classic KG embeddings such as RDF2Vec \cite{ristoski2016rdf2vec}, TransE \cite{bordes2013translating} and DistMult \cite{yang2015embedding}, \mr{and two supervised Transformer \cite{vaswani2017attention} classifiers based on the textual context.}
We also calculated the Euclidean distance between entities and visualized the embeddings of some example entities to analyze different embedding methods. 

\mr{
Briefly this study can be summarized as follows.
\begin{itemize}
  \item This work is among the first that aim at embedding all kinds of semantics of OWL ontologies including the graph structure, the literals and the logical constructors. A general framework named \owlvec together with different strategies for addressing different OWL semantics has been developed. It would \textit{(i)} enable many statistical or machine learning tasks over massive ontologies, thus assisting their curation and boosting their application, \textit{(ii)} provide a potential direction for bridging symbolic and sub-symbolic knowledge representation for new neural-symbolic solutions.
  \item The work has evaluated \owlvec for two important ontology completion case studies (class membership prediction and subsumption prediction) on three real world ontologies, where \owlvec outperforms the current ontology embedding methods and the classic KG embedding methods. It has also conducted extensive ablation studies to verify the adopted strategies as well as visualization analysis for interpretation.
\end{itemize}
}

The remainder of the paper is organized as follows. 
The next section introduces the preliminaries including both background and related work.
Section 3 introduces the technical details of \owlvec as well as the case studies.
Section 4 presents the experiments and the evaluation results.
The last section concludes and discusses future work.

\section{Preliminaries}

\subsection{\owl Ontologies}
Our \owlvec embedding targets \owl ontologies \cite{bechhofer2004owl}, which are based on the \sroiq{} description logic (DL) \cite{baader2017introduction}.
Consider a signature $\Sigma=(\mathcal{N}_C, \mathcal{N}_R, \mathcal{N}_I)$, where $\mathcal{N}_C$, $\mathcal{N}_R$ and $\mathcal{N}_I$ are pairwise disjoint sets of,
respectively, atomic concepts, atomic roles  and individuals. 
Complex concepts and roles can be composed using DL constructors such as 
conjunction (e.g., $C \sqcap D)$, disjunction (e.g., $C \sqcup D$), existential restriction (e.g., $\exists r.C$) and universal restrictions (e.g., $\forall r.C$) where $C$ and $D$ are concepts, and $r$ is a role.
An \owl ontology comprises a TBox $\mathcal{T}$ and an ABox $\mathcal{A}$.
The TBox is a set of axioms such as 
General Concept Inclusion (GCI) axioms (e.g., $C \sqsubseteq D$), Role Inclusion (RI) axioms 
(e.g., $r \sqsubseteq s$) and Inverse Role axioms (e.g., $s \equiv r^-$), where $C$ and $D$ are concepts, $r$ and $s$ are roles, and $r^-$ denotes the inverse of $r$.
The ABox is a set of assertions such as concept assertions (e.g., $C(a)$), role assertions (e.g., $r(a, b)$) and individual equality and inequality assertions (e.g., $a \equiv b$ and $a \not\equiv b$), where $C$ is a concept, $r$ is a role, $a$ and $b$ are individuals.
\mr{It is worth noting that the terminological part can also be divided into a TBox and an RBox, where the RBox models the interdependencies between the roles such as the RI.}

\begin{figure}[t]
\begin{center}
\begin{subfigure}[b]{\textwidth}
\centering
\includegraphics[width=0.8\textwidth]{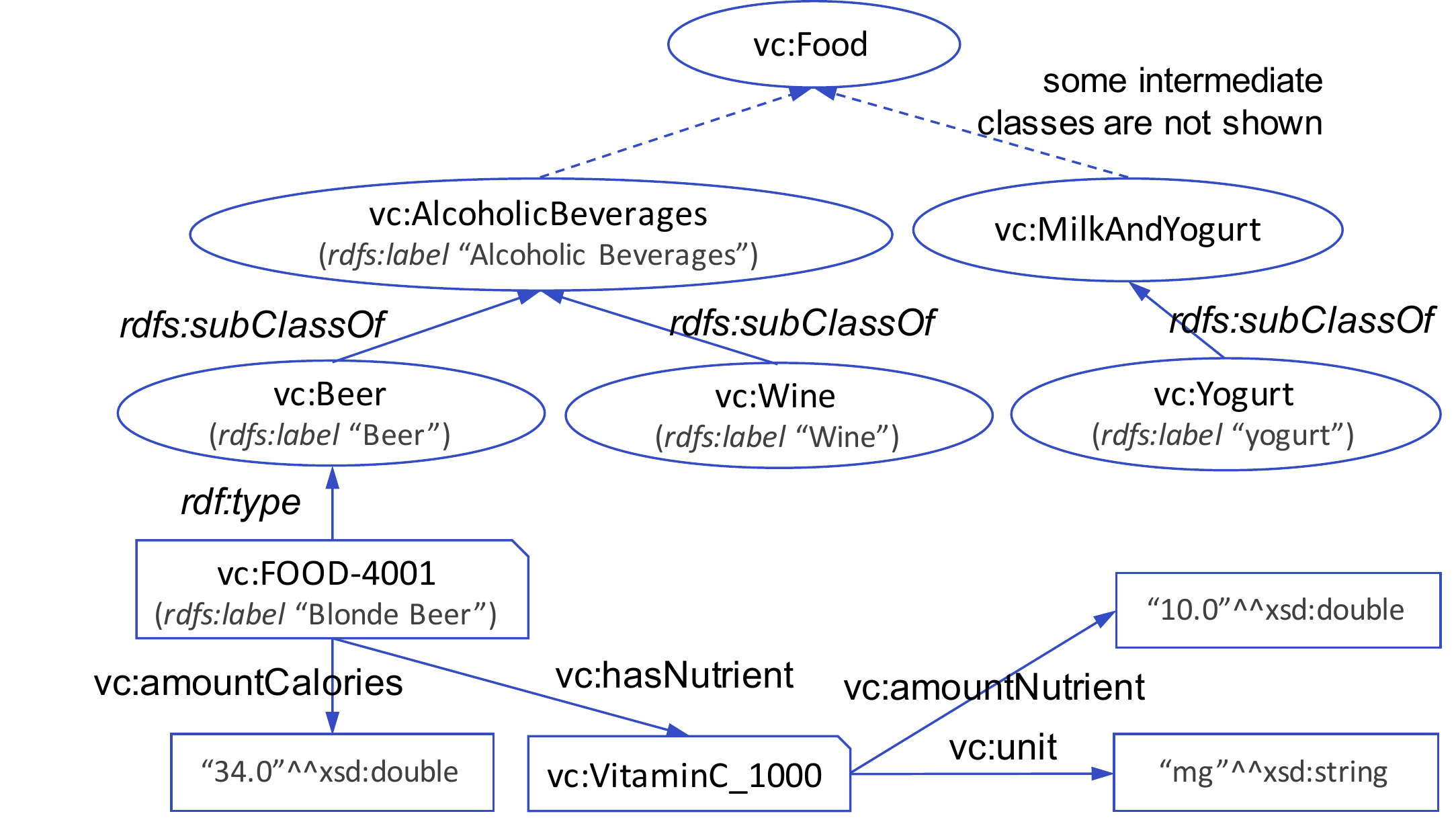}
\vspace{-0.1cm}
\caption{The HeLis Ontology}\label{fig:helis_example}
\end{subfigure}
\hfill
\vspace{0.1cm}
\begin{subfigure}[b]{\textwidth}
\centering
\includegraphics[width=0.8\textwidth]{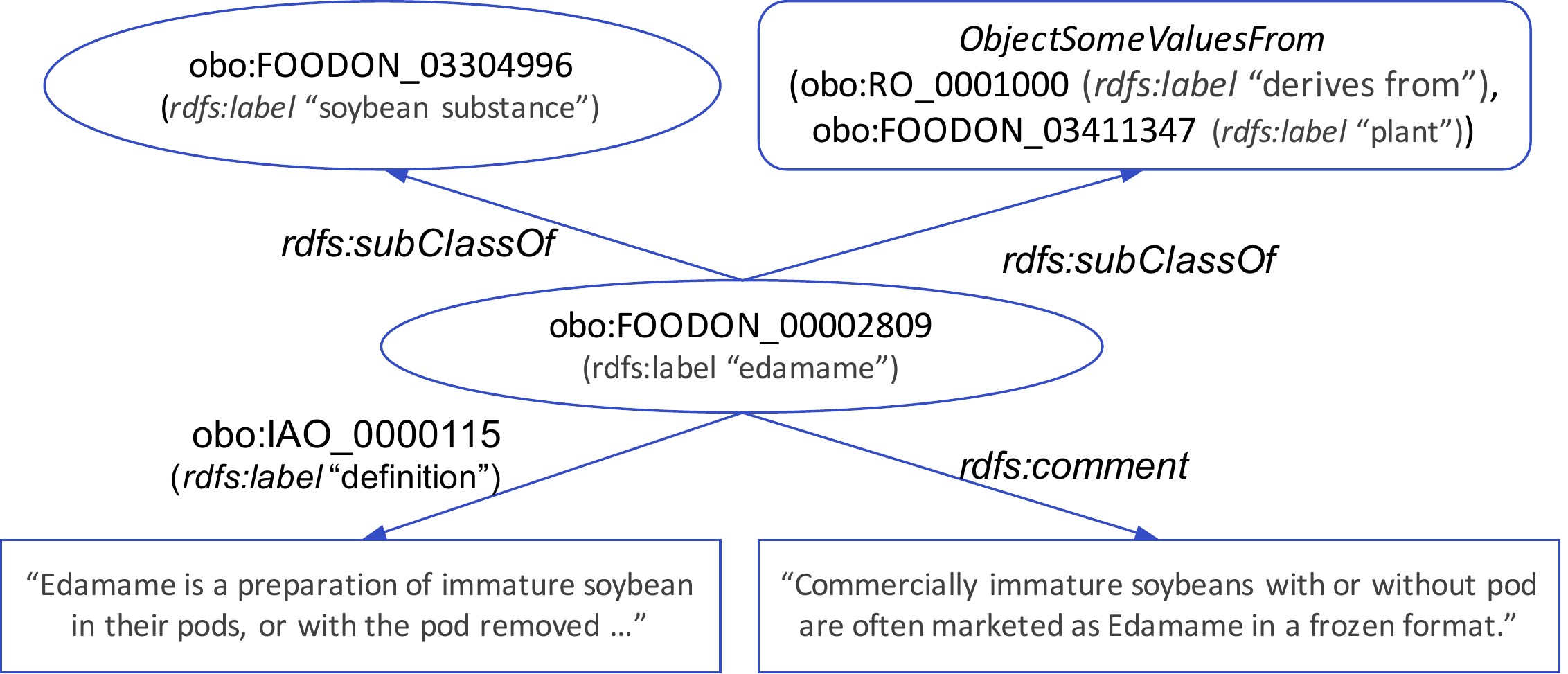}
\vspace{-0.1cm}
\caption{The FoodOn Ontology}\label{fig:foodon_example}
\end{subfigure}
\caption{\small \mr{Fragments of the ontologies.}\protect\footnotemark}\label{fig:examples}
\end{center}
\end{figure}

\mr{
In \owl, the aforementioned concept, role and individual are modeled as \textit{class}, \textit{object property} and \textit{instance}, respectively.
To avoid confusion, we will only use the terms of class, object property and instance in the remainder of the paper.
We will also use the term \textit{property} which can refer to not only object property, but also \textit{data property} and \textit{annotation property}.
Meanwhile, for convenience, we will also use a general term \textit{entity} to refer to a class, a property or an instance.
Note object property models the relationship between two instances, data property models the relationship between an instance and a literal value (e.g., number and text), and annotation property models the relationship between an entity and an annotation (e.g., comment and label).
Each entity is uniquely represented by an Internationalized Resource Identifier (IRI)\footnote{An entity can also be represented by an Uniform Resource Identifier (URI). IRIs extend URIs by using the Universal Character Set, where URIs were limited to ASCII, with far fewer characters.}.
These IRIs may be lexically `meaningful' (\eg \textit{vc:AlcoholicBeverages} in Figure~\ref{fig:helis_example}) or consist of internal IDs that do not carry useful lexical information (\eg \textit{obo:FOODON\_00002809} in Figure~\ref{fig:foodon_example}); in either case the intended meaning may also be
indicated via annotations (see below).
To make it easier to follow, we will append the ID-based IRI by a readable name from its label when necessary.

In \owl, a GCI axiom $C \sqsubseteq D$ corresponds to a subsumption relation between the class $C$ and the class $D$, while a concept assertion $C(a)$ corresponds to a membership relation between the instance $a$ and the class $C$.
Meanwhile, in \owl, complex concepts, complex roles, axioms and role assertions can be serialised as (sets of) RDF triples, each of which is a tuple composed of a subject, a predicate and an object. 
For the predicate, these triples use a combination of bespoke object properties (e.g., \textit{vc:hasNutrient}), and built-in properties by RDF, RDFS\footnote{\url{https://www.w3.org/TR/rdf-schema/}} and \owl (e.g., \textit{rdfs:subClassOf}, \textit{rdf:type} and \textit{owl:someValuesFrom}).}
In Fig.~\ref{fig:examples}, for example, the relationship between the two instances \textit{vc:FOOD-4001 \mr{(blonde beer)}} and \textit{vc:VitaminC\_100} is represented by 
a triple using the object property \textit{vc:hasNutrient}, while the existential restriction involving the class 
\textit{obo:FOODON\_00002809 \mr{(edamame)}} and the object property \textit{obo:RO\_0001000 \mr{(derives from)}} is represented by triples using three \owl built-in properties, i.e., \textit{owl:Restriction}, \textit{owl:onProperty} and \textit{owl:someValuesFrom}.
\mr{The object of an RDF triple of an \owl assertion can be a literal value;
for example, the calories amount of \textit{vc:FOOD-4001 \mr{(blonde beer)}} is represented by a triple using the bespoke data property 
\textit{vc:amountCalories} and the literal value $34.0$ of type \textit{xsd:double}. }

In addition to axioms and assertions with formal logic-based semantics, 
an ontology often contains \ejr{metadata information in the form of annotation axioms}.
\mr{These annotations can also be represented by RDF triples
using annotation properties as the predicates}; e.g., the class \textit{obo:FOODON\_00002809 \mr{(edamame)}} is annotated using \textit{rdfs:label} to specify a name string, using \textit{rdfs:comment} to specify a description, and using \textit{obo:IAO-0000115 \mr{(definition)}} --- a bespoke annotation property to specify a natural language ``definition''.

\footnotetext{\label{ft:prefix}
Note \textit{vc} is the prefix associated to the IRI namespace of \url{http://www.fbk.eu/ontologies/virtualcoach\#}, while \textit{obo}, \textit{oboInOwl}, \textit{xsd}, \textit{rdf}, \textit{rdfs} and \textit{owl} are prefixes of standard vocabularies. }

Knowledge graph (KG) refers to structured knowledge resources which are often expressed as a set of RDF triples \cite{hogan2020knowledge}.
Many KGs only contain instances and facts which are equivalent 
to an OWL ontology ABox.
Some other KGs such as DBpedia \cite{auer2007dbpedia} 
are also enhanced with an schema which is equivalent to the TBox of an OWL ontology. Thus, a KG can often be understood as an ontology.

\subsection{Semantic Embedding}
Semantic embedding refers to a series of representation learning (or feature learning) techniques that encode the semantics of data such as sequences and graphs into vectors, such that they can be utilized by downstream machine learning prediction and statistical analysis tasks \cite{bengio2013representation}. 
\mr{Word embedding or sequence feature learning} models such as Feed-Forward Neural Networks, Recurrent Neural Networks and Transformers are widely used for semantic embedding, and they have shown good performance in embedding the context (e.g., item co-occurrence) in sequences \cite{mikolov2013efficient,peters2018deep,devlin2019bert}.
Two classic auto-encoding architectures for learning representations of sequential items are continuous skip-gram and continuous Bag-of-Words (CBOW) \cite{mikolov2013distributed,mikolov2013efficient}.
The former aims at predicting the surroundings of an item, while the latter aims at predicting an item based on its surroundings.
\textit{Word2Vec} is a well known group of \mr{sequence feature learning techniques} for learning word embeddings from a large corpus, and was initially developed by a team at Google; it can be configured to use either skip-gram or CBOW architectures 
\cite{mikolov2013distributed,mikolov2013efficient}.

Semantic embedding has also been extended to KGs composed of role assertions \cite{wang2017knowledge}.
The entities and relations (object properties) are represented in a vector space 
while retaining their relative relationships (semantics), 
and the resulting vectors are then applied to downstream tasks including link prediction \cite{rossi2020knowledge}, entity alignment \cite{sun2020benchmarking}, and erroneous fact detection and correction \cite{chen2020correcting}.
One paradigm for learning KG representations is computing the embeddings in an \textit{end-to-end} manner, iteratively adjusting the vectors using an optimization algorithm to minimize the overall loss across all the triples,
where the loss is usually calculated by scoring the truth/falsity of each triple 
(positive and negative samples).
Algorithms based on this technique include translation based models such as TransE~\cite{bordes2013translating} and TransR~\cite{lin2015learning} and latent factor models such as DistMult~\cite{yang2015embedding}.

Another paradigm is to first explicitly explore the neighborhoods of entities and relations in the graph, and then learn the embeddings using a \mr{word embedding} model.
Two representative algorithms based on this paradigm are node2vec \cite{grover2016node2vec} and Deep Graph Kernels \cite{yanardag2015deep}.
The former extracts random graph walks and creates skip-gram or CBOW models as the corpus for training,
while the latter uses graph kernels such as Weisfeiler-Lehman (WL) sub-graph kernels as the corpus.
However, both embedding algorithms were originally developed for undirected graphs,
and thus may have limited performance when directly applied to KGs.
RDF2Vec addresses this issue by extending the idea of the above two algorithms to directed labeled RDF graphs, and has been shown to learn effective embeddings for large scale KGs such as DBpedia \cite{ristoski2016rdf2vec,ristoski2019rdf2vec}.
Recent studies have explored the usage of new \mr{word embedding or sequence feature learning models for learning embeddings; one example is the KG embedding and link prediction method named RW-LMLM which combines the random walk algorithm with Transformer \cite{wang2019capturing}.}

Our \owlvec technique belongs to the \mr{word embedding} paradigm, but we focus on \owl ontologies instead of typical KGs, with the goal of preserving the semantics 
not only of the graph structure, but also of the lexical information and the logical constructors.
Note that the graph of an ontology, which includes hierarchical categorization structure, differs from the multi-relation graph composed of \mr{role (relation) assertions} of a typical KG;
\mr{furthermore, according to our literal review on ontology embedding (cf. Section \ref{sec:oe}) and the latest survey \cite{kulmanov2020semantic}, there are currently no existing KG embedding methods that jointly explore the ontology's lexical information and logical constructors.}

\subsection{Ontology Embedding}\label{sec:oe}
The use of machine learning prediction and statistical analysis with ontologies is receiving wider attention, and some approaches to embedding the semantics of \owl ontologies can already be found in the literature.
Unlike 
\ejr{typical} KGs, \owl ontologies include not only graph structure but also logical constructors, 
and entities are often augmented with richer lexical information specified using \textit{rdfs:label}, \textit{rdfs:comment} and many other bespoke or built-in annotation properties.
The objective of \owl ontology embedding in this study is to represent each \owl named entity (class, instance or property) by a vector, such that the inter-entity relationships indicated by the above information are kept in the vector space, and the performance of the downstream tasks, where the input vectors can be understood as learned features, is maximized.

EL Embedding \cite{kulmanov2019embeddings} and Quantum Embedding \cite{garg2019quantum} are two \owl ontology embedding algorithms of the end-to-end paradigm.
They construct specific score functions and loss functions for logical axioms from $\mathcal{EL}^{++}$ and $\mathcal{ALC}$, respectively, by transforming logical relations into geometric relations.
This encodes the semantics of the logical constructors, but ignores the additional semantics provided by the lexical information of the ontology.
Moreover, although the graph structure is explored by considering class subsumption and class membership axioms, the exploration is incomplete as it uses only \textit{rdfs:subClassOf} and \textit{rdf:type} edges, and ignores edges involving other relations.

Onto2Vec \cite{smaili2018onto2vec} and OPA2Vec \cite{smaili2018opa2vec} are two ontology embedding algorithms of the \mr{word embedding} paradigm using a model of either the skip-gram architecture or the CBOW architecture.
Onto2Vec uses the axioms of an ontology as the corpus for training,
while OPA2Vec complements the corpus of Onto2Vec with the lexical information provided by, e.g., \textit{rdfs:comment}. 
Both adopt the deductive closure of an ontology with entailment reasoning.
They have been evaluated with the Gene Ontology for predicting protein-protein interaction (i.e., a domain-specific relationship between classes), which is quite different from the class membership prediction and the class subsumption prediction in this study.
Both methods treat each axiom as a sentence, which means that they cannot explore the correlation between axioms.
This makes it hard to fully explore the graph structure and the logical relation between axioms, and may also lead to the problem of corpus shortage for small to medium scale ontologies.
\owlvec deals with the above issues of OPA2Vec and Onto2Vec by complementing their axiom corpus with a corpus generated by walking over RDF graphs that are transformed from the \owl ontology with its graph structure and logical constructors considered.
In addition, to fully utilize the lexical information, 
\owlvec creates embeddings for not only the ontology entities as the current KG/ontology embedding methods but also for the words in the lexical information.

%

\vspace{-0.2cm}

\section{Methodology}
\begin{figure}[t]
\begin{center}
\includegraphics[width=0.88\textwidth]{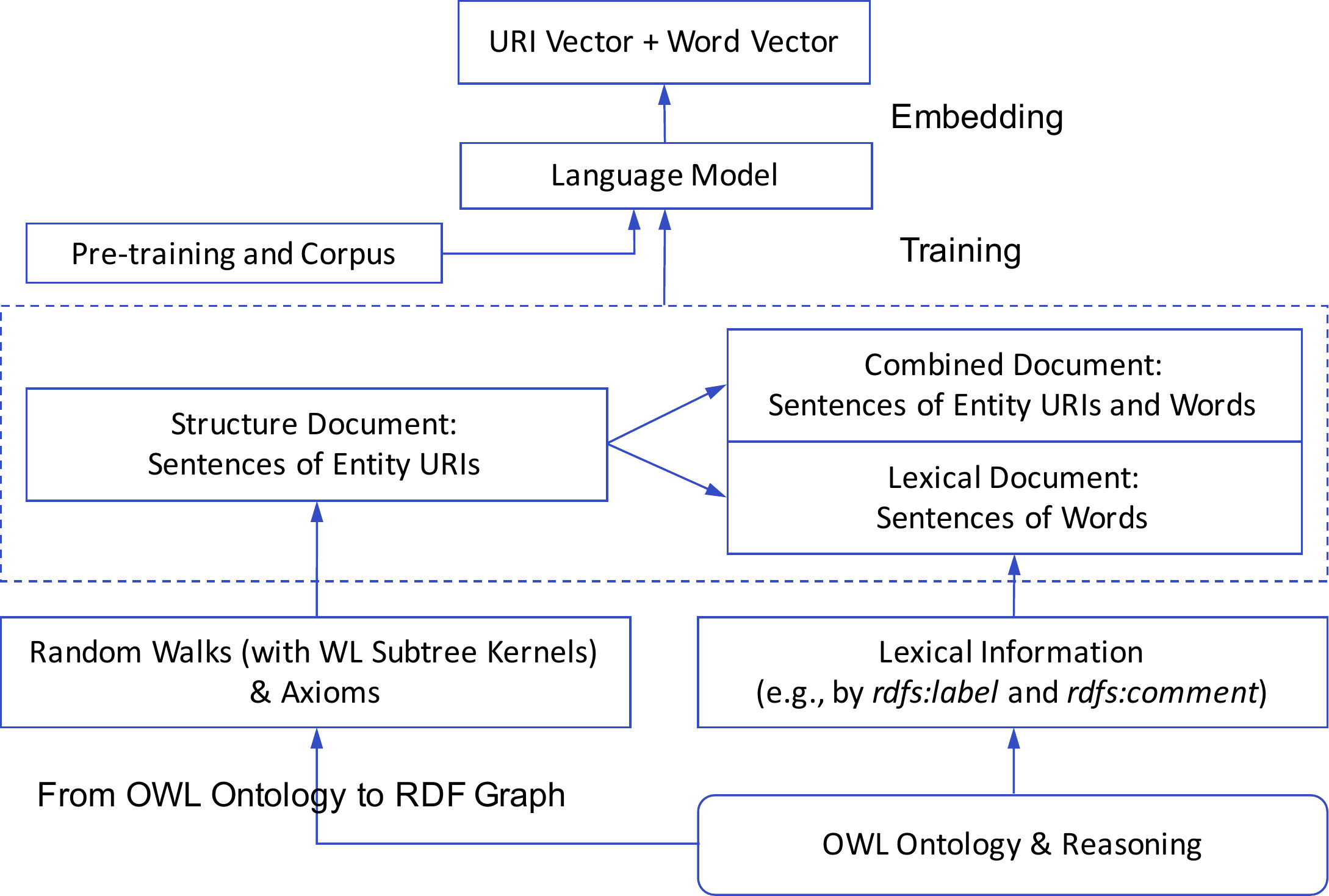}
\caption{\footnotesize The overall framework of \owlvec.  }
\label{fig:framework}
\end{center}
\end{figure}

Figure \ref{fig:framework} presents the overall framework of \owlvec, which mainly consists of two core steps: \textit{(i)} corpus extraction from the ontology, and \textit{(ii)} \mr{word embedding} model training with the corpus.
\mr{The corpus includes a structure document, a lexical document, and a combined document.}
The first two documents aim at exploring the ontology's graph structure, logical constructors and lexical information, \mr{where ontology entailment reasoning can be enabled,} 
while the third document aims at preserving the correlation between entities (IRIs) and their lexical labels (words).
\mr{Note the latter two documents are constructed using the first document as the backbone together with kinds of lexical information of the ontology.
See Table \ref{tab:sentenceExample} for sentence examples of each document.}
Briefly, given an input ontology $\mathcal{O}$ and the target entities $E$ of $\mathcal{O}$ for embedding, \owlvec outputs a vector for each entity $e$ in $E$, denoted as $\bm{e} \in \mathbb{R}^d$, where $d$ is the (configurable) embedding dimension.
Note that $E$ can be all the entities in $\mathcal{O}$ or just a part needed for a specific application. 
\mr{With the \owlvec embeddings, we apply them in two downstream case studies --- class membership prediction and class subsumption prediction.}
For class membership prediction we set $E$ to all the named classes and instances; for class subsumption prediction we set $E$ to all the named classes. 



\subsection{From \owl Ontology to RDF Graph}\label{sec:otog}

\setlength{\tabcolsep}{3.0pt}
\begin{table}[t] 
\renewcommand{\arraystretch}{1.3}
\centering 
    {\scriptsize
    \begin{tabular}{p{2.8cm}<{\centering}|p{5cm}<{\centering}|p{3.4cm}<{\centering}}\hline
    
    
    \multicolumn{1}{c|}{\textbf{Axiom of Condition 1}}    &    \multicolumn{1}{c|}{\textbf{Axiom or Triple(s) of Condition 2}}   &    \multicolumn{1}{c}{\textbf{Projected Triple(s)}} \\ \hline

    $A \sqsubseteq \Box r.D$  & 
    \multirow{3}{*}{$D\equiv B ~|~ B_1 \sqcup...\sqcup B_n ~|~ B_1
    \sqcap...\sqcap B_n$} & \multirow{5}{*}{$\langle A, r, B \rangle$ or} \\
    or & &   \\
    $\Box r.D \sqsubseteq A$  &  &     \\\cline{1-2}
    $\exists r.\top \sqsubseteq A$ (domain)    &    $\top \sqsubseteq \forall r.B$  (range)   & $\langle A, r, B_i \rangle$ for $i \in 1,...,n$     \\\cline{1-2}
    $A \sqsubseteq \exists r.\{b\}$  &    $B(b)$  & \\\cline{1-2}
    $r \sqsubseteq r^{\prime}$    &  
    $\langle A, r^{\prime}, B \rangle$ has been projected   
    &  \\\cline{1-2}
    $r^{\prime} \equiv r^{-}$    &  
    $\langle B, r^{\prime}, A \rangle$ has been projected   
    &  \\\cline{1-2}
    $s_1\ \circ\ ...\ \circ \ s_n  \sqsubseteq r $    &  $\langle A,
    s_1, C_1 \rangle$...$\langle C_n, s_n, B \rangle$ have been projected   &  \\\hline




    \multirow{2}{*}{$B \sqsubseteq A$} &
    \multirow{2}{*}{--} & $\langle B, rdfs\nsp:\nsp subClassOf, A \rangle$ \\ 
    &  & $\langle A, rdfs\nsp:\nsp subClassOf^{-}, B \rangle$\\ \hline
    
    \multirow{2}{*}{$A(a)$} & \multirow{2}{*}{--} &  $\langle a, rdf\nsp:\nsp type, A \rangle$\\
    & & $\langle A, rdf\nsp:\nsp type^{-}, a \rangle$\\\hline
    $r(a, b)$ & -- &  $\langle a, r, b \rangle$\\\hline

\end{tabular}
}
\caption{\footnotesize \ejr{Projection rules, based on  \cite{optiquevqs2018,holter2019embedding}, used in the second strategy to generate an RDF graph. $\Box$ is one of: $\geq$, $\leq$, $=$, $\exists$,~$\forall$. $A$, $B$, $B_i$ and $C_i$ are atomic concepts (classes), $s_i$, $r$ and $r^{\prime}$ are roles (object properties), $r^{-}$ is the inverse of a relation $r$, $a$ and $b$ are individuals (instances), $\top$ is the top concept (defined by \textit{owl:Thing}).}}
\label{tab:ontologyProjection}
\end{table}

\owlvec incorporates two strategies to turn the  original \owl ontology $\mathcal{O}$ into a graph $\mathcal{G}$ that is composed of RDF triples. 
The first strategy is the transformation according to \owl to RDF Graph Mapping \mr{which is originally defined by W3C\footnote{\label{fn1-mapping}\url{https://www.w3.org/TR/owl2-mapping-to-rdf/}} to store and exchange OWL ontologies by RDF triples.
Some simple axioms such as membership and subsumption axioms for atomic entities, data and annotation properties associated to atomic entities, and relational assertions between atomic instances can be directly transformed into RDF triples by introducing some built-in properties or using the bespoke properties in the axioms (e.g., $\langle$\textit{vc:FOOD-4001 (Blonde Beer)}, \textit{rdf:type}, \textit{vc:Beer}$\rangle$, $\langle$\textit{vc:FOOD-4001}, \textit{rdfs:label}, ``Blonde Beer''$\rangle$ and $\langle$\textit{vc:FOOD-4001}, \textit{vc:hasNutrient}, \textit{vc:VitaminC\_1000}$\rangle$).
Some complex axioms such as some logical constructors and those axioms involving complex entities need to be transformed into multiple triples and often rely on black nodes.}
For example, the existential restriction of the class \textit{obo:FOODON\_00002809 \mr{(edamame)}} in Figure \ref{fig:foodon_example}, i.e., \textit{ObjectSomeValuesFrom(obo:RO\_0001000 \mr{(derives from)},  obo:FOODON\_03411347 \mr{(plant)})} is transformed into four RDF triples, i.e., $\langle$\textit{obo:FOODON\_00002809}, \textit{rdfs:subClassOf}, \textit{\_:x}$\rangle$,  $\langle$\textit{\_:x}, \textit{owl:someValuesFrom},  \textit{obo:FOODON\_03411347} $\rangle$, $\langle$\textit{\_:x}, \textit{rdf:type}, \textit{owl:Restriction}$\rangle$ and $\langle$\textit{\_:x}, \textit{owl:onProperty}, \textit{obo:RO\_0001000}  $\rangle$, where \textit{\_:x} denotes a blank node.
\mr{In this example, one additional node \textit{\_:x} and one additional edge \textit{rdfs:subClassOf} are inserted between \textit{obo:FOODON\_00002809} and \textit{obo:FOODON\_03411347}.}

The second strategy is based on projection rules proposed in \cite{optiquevqs2018,holter2019embedding}, as shown in Table \ref{tab:ontologyProjection}, \mr{where
every RDF triple $\langle X, r, Y \rangle$ in the projection (the third column) is justified by one or more axioms in the ontology (the first and second columns). 
As in the first strategy, a simple relational assertion between two atomic entities (the final row in Table \ref{tab:ontologyProjection}), or a simple data or annotation property associated to an atomic entity, is directly transformed into one single triple.
While those complex logical constructors (the first six rows in Table \ref{tab:ontologyProjection}), unlike the first strategy, are approximated.}
For example, the above mentioned existential restriction of the class \textit{obo:FOODON\_00002809} would be represented with $\langle$\textit{obo:FOODON\_00002809}, \textit{obo:RO\_0001000},  \textit{obo:FOODON\_03411347}$\rangle$. 
This avoids the use of blank nodes in the RDF graph, which may act as noise towards the correlation between entities when the embeddings are learned; but, not the exact logical relationships are kept in the resulting RDF graph.
\mr{Specially, the projection of membership and subsumption axioms (the seventh and eighth rows in Table \ref{tab:ontologyProjection}) includes two settings.
In the first setting, the two involved atomic entities are transformed into one triple with the predicate of \textit{rdf:type} or \textit{rdfs:subClassOf}.
In the second setting, besides the above triple, one more triple which uses the inverse of \textit{rdf:type} or \textit{rdfs:subClassOf} is added.
This enables a bidirectional walk between two entities with a subsumption or membership relationship on the transformed RDF graph, and would impact the corpus and the embeddings.
In the remainder of this paper, we by default refer to the first setting when we mention projection rules, and we refer to the second setting by the term of projection rules with inverse or the appended annotation of (+R).}

Both ontology to RDF graph transformation strategies can incorporate an \owl entailment reasoner to compute the TBox classification and ABox realization before $\mathcal{O}$ is transformed into an RDF graph $\mathcal{G}$. 
Such reasoning grounds the axioms of logical constructors and leads to explicit representation of some hidden knowledge. 
\mr{For example, in Fig. \ref{fig:examples}(a), we can infer a hidden triple $\langle$\textit{vc:FOOD-4001 (blonde beer)}, \textit{rdf:type}, \textit{vc:AlcoholicBeverages}$\rangle$ from $\langle$\textit{vc:FOOD-4001}, \textit{rdf:type}, \textit{vc:Beer}$\rangle$ and $\langle$\textit{vc:Beer}, \textit{rdfs:subClassOf}, \textit{vc:AlcoholicBeverages}$\rangle$.
When the reasoning is enabled, such inferred hidden triples will be included in the transformed RDF graph $\mathcal{G}$.
}
In our experiments we use the HermiT OWL reasoner \cite{glimm2014hermit}, and we evaluate the impact of enabling or disabling reasoning \mr{(cf. the second paragraph in Section \ref{sec:lc} and Table \ref{res:inference}).}

\subsection{Structure Document}

\setlength{\tabcolsep}{3.0pt}
\begin{table}[t] 
\renewcommand{\arraystretch}{1.3}
\centering 
{\scriptsize
\begin{tabular}{p{1.3cm}<{\centering}|m{4.2cm}<{\centering}|m{5.7cm}<{\centering}}\hline
    
\tabincell{\textbf{Document}} & \tabincell{\textbf{Example Name and Source}} & \tabincell{\textbf{Sentence(s)}} \\ \hline

\multirow{4}{*}{\tabincell{\\ \\ Structure}} & Ex1: random walk of depth tree starting from \textit{vc:FOOD-4001 (blonde beer)} & (\textit{vc:FOOD-4001}, \textit{vc:hasNutrient}, \textit{vc:VitaminC\_100}, \textit{vc:amountNutrient})    \\ \cline{2-3}
&  Ex2: random walk of depth four starting from \textit{vc:FOOD-4001 (blonde beer)} & (\textit{vc:FOOD-4001}, \textit{rdf:type}, \textit{vc:Beer}, \textit{rdfs:subClassOf}, \textit{vc:AlcoholicBeverages}) \\ \cline{2-3}
&Ex3: the walk in Ex2 with the WL sub-graph kernel enabled  & (\textit{vc:FOOD-4001}, \textit{rdf:type}, \textit{kernel\_id1\_md5}, \textit{rdfs:subClassOf}, \textit{kernel\_id2\_md5}) \\ \cline{2-3}
&Ex4: the existential restriction of \textit{obo:FOODON\_00002809 (edamame)} & (\textit{obo:FOODON\_00002809}, \textit{subClassOf}, \textit{obo:RO\_0001000}, some, \textit{obo:FOODON\_03411347}) \\ \hline

\multirow{3}{*}{\tabincell{\\ Literal}} &Ex5: replacing the IRIs in Ex1 by their labels or names & (``blonde'', ``beer'', ``has'', ``nutrient'', ``vitamin'', ``c'', ``amount'', ``nutrient'') \\ \cline{2-3}
&Ex6: replacing the IRIs in Ex3 by their labels or names & (``blonde'', ``beer'', ``type'', kernel\_id1\_md5, ``subclassof'', kernel\_id2\_md5) \\ \cline{2-3}
&Ex7: the \textit{obo:IAO\_0000115 (definition)}  annotation of \textit{obo:FOODON\_00002809 (edamame)}  & (``edamame'', ``edamame'', ``is'', ``a'', ``preparation'', ``of'', ``immature'', ``soybean'', ``in'', ``their'', ``pods'' ...) \\ \hline

\multirow{2}{*}{\tabincell{\\ Combined}} & Ex8: keeping the IRI of one item in the Ex1 sentence (random strategy) & (\textit{vc:FOOD-4001}, ``has'', ``nutrient'', ``vitamin'', ``c'', ``amount'', ``nutrient'') \\ \cline{2-3}
&Ex9: keeping the IRI of one item in the Ex1 sentence (traversing strategy) & (\textit{vc:FOOD-4001}, ``has'', ``nutrient'', ``vitamin'', ``c'', ``amount'', ``nutrient''), $\dots$, (``blonde'', ``beer'', ``has'', ``nutrient'', ``vitamin'', ``c'', \textit{vc:amountNutrient}) \\ \hline
    
\end{tabular}
}
\caption{\footnotesize \mr{Sentence examples that are extracted from the ontology fragments in Fig. \ref{fig:examples}.}}
\label{tab:sentenceExample}
\end{table}

The structure document aims at capturing both the graph structure and the logical constructors of the ontology. 
With the RDF graph $\mathcal{G}$, one option is computing  
random walks for each target entity in $E$ with the RDF graph $\mathcal{G}$. 
Each walk, which is a sequence of entity IRIs, acts as a sentence of the structure document.
\mr{
Ex1 and Ex2 in Table \ref{tab:sentenceExample} are two walk examples both starting from the class \textit{vc:FOOD-4001 (blonde beer)}.
To implement the random walk algorithm, we first transform the RDF graph $\mathcal{G}$ into a directed single relation graph $\mathcal{G}'$; for each RDF triple $\langle X, r, Y \rangle$ in $\mathcal{G}$, the subject $X$, the object $Y$ and the relation $r$ are transformed into three vertices, two edges are added from the vertex of $X$ to the vertex of $r$ and from the vertex of $r$ to the vertex of $Y$ respectively.
Given one starting vertex, we fairly and randomly select the next vertex from all its connected vertices, and iterate this ``step'' operation for a specific number of times as the procedure of ``walking''.
%

\owlvec also allows the usage of the Weisfeiler Lehman (WL) kernel \cite{shervashidze2011weisfeiler} which encodes the structure of a sub-graph into a unique identity and thus enables the representation and incorporation of the sub-graph in a walk.
For one vertex in the transformed single relation graph $\mathcal{G}'$, there is an associated sub-graph (neighbourhood) starting from this vertex, and we simply call this sub-graph's WL kernel (identity) as this vertex's WL kernel.
In implementation, we first extract the original random walks.
For each random walk, we then keep the IRIs of the starting vertex and the vertices that are transformed by the relations, but replace the IRIs of the none-starting vertices that are transformed by the subjects or objects with their WL kernels.
Ex3 in Table \ref{tab:sentenceExample} is an example of enabling the WL sub-graph kernel for the walk of Ex2.
Note in calculating a vertex's WL kernel, the size of its sub-graph, i.e., the depth from this vertex to the farthest vertex in the sub-graph can be set. 
We generate and adopt all the walks, with the sub-graph size ranging from $0$ to a maximum size --- a hyper-parameter that is set to $4$ by default.
Specially, the WL kernel enabled random walk with the sub-graph size of $0$ is equivalent to the original random walk.
}

To capture the logical constructors, \owlvec extracts all the axioms of the ontology as a complement of the sentences of the structure document.
\mr{
In implementation, each ontology axiom is transformed into a sequence following the \owl Manchester Syntax\footnote{\url{https://www.w3.org/TR/owl2-manchester-syntax/}}, where the original built-in terms such as ``subClassOf'' and ``some'' are kept.
Ex4 in Table \ref{tab:sentenceExample} is an example of such Manchester Syntax sentence according to the axiom of the existential restriction of the class \textit{obo:FOODON\_00002809 (edamame)}.
In comparison with the random walk over the projected RDF graph, which generates the sentence of (\textit{obo:FOODON\_00002809}, \textit{obo:RO\_0001000}, \textit{obo:FOODON\_03411347}) for the same axiom in Ex4, the Manchester Syntax sentence indicates the logical relationship between the terms by the buit-in terms;
while in comparison with the random walk over the graph transformed by W3C OWL to RDF Graph Mapping, the Manchester syntax sentence is shorter and avoids the blank nodes.
}


\subsection{Lexical Document}
\mr{The lexical document includes two kinds of word sentences.
The first kind are generated from the entity IRI sentences in the structure document,
while the second are extracted from the relevant lexical annotation axioms in the ontology.
For the first kind,} given an entity IRI sentence, each of its entities is replaced by its English label defined by \textit{rdfs:label}.
Note the label is parsed and transformed into lowercase tokens, and those tokens with none letter characters are filtered out, before it replaces the entity IRI. 
It is possible that some entities have no English annotations by \textit{rdfs:label}, 
such as the class \textit{vc:MilkAndYogurt} and the instance \textit{vc:VitaminC\_1000} in Figure \ref{fig:helis_example}.
In this case, we prefer to use the name part of the IRI \mr{and parse it into words assuming that the name follows the camel case (e.g., \textit{vc:MilkAndYogurt} is parsed into ``milk'', ``and'' and ``yogurt'').
One sentence example of this kind is Ex5 in Table \ref{tab:sentenceExample} which is generated by replacing the IRIs of the Ex1 sentence by their words.
Specially, some IRIs have neither English labels or meaningful IRI names, and when the WL sub-graph kernel is enabled, there are also kernel identities in the structure sentence. 
We keep these original IRIs and identities in the word sentences (cf. Ex6 in Table \ref{tab:sentenceExample}).
}

\mr{The second kind of word sentences are extracted from those textual annotations. They include two kinds: annotations by bespoke annotation properties such as \textit{obo:IAO\_0000115 (definition)}, \textit{obo:IAO\_0010000 (has axiom label)} and \textit{oboInOwl:hasSynonym}\footref{ft:prefix},
and annotations by built-in annotation properties such as \textit{rdfs:comment} and \textit{rdfs:seeAlso}.
In our current \owlvec implementation, we consider all the annotation properties of an ontology except for \textit{rdfs:label}.
The annotations by \textit{rdfs:label} are ignored in generating word sentences of the second kind because they are already considered in the word sentences of the first kind (e.g., Ex5).
}
More specifically, for each  annotation axiom, \owlvec replaces the subject entity by its English label or IRI name as in transforming the IRI sentence, and keeps the lowercase word tokens parsed from the annotation value.
\mr{One example of such word sentence is Ex7 in Table \ref{tab:sentenceExample} which is based on the annotation by \textit{obo:IAO\_0000115 (definition)} to the class \textit{obo:FOODON\_00002809 (edamame)}.
It would enable the model to learn the correlation of ``edamame'' to other words in the relevant background such as ``soybean'' and ``pods''.
}

\subsection{Combined Document}
\owlvec further extracts a combined document from the structure document and the entity annotations, so as to preserve the correlation between entities (IRIs) and words in the lexical information.
To this end, we developed two strategies to deal with each IRI sentence in the structure document.
One strategy is to randomly select an entity in an IRI sentence, keep the IRI of this entity, and replace the other entities of this sentence by their lowercase word tokens extracted from their labels or IRI names as in the creation of the lexical document.
\mr{One example is Ex8 in Table \ref{tab:sentenceExample}, where the IRI of \textit{vc:FOOD-4001 (blonde beer)} of the IRI sentence of Ex1 is kept while the other IRIs are replaced by their corresponding words.}
The other strategy is traversing all the entities in a IRI sentence.
For each entity, it generates a combined sentence by keeping the IRI of this entity, and replacing the others by their lowercase word tokens as in the random strategy.
Thus for one IRI sentence, it generates $m$ combined sentences where $m$ is the number of entities of the IRI sentence.
\mr{Ex9 in Table \ref{tab:sentenceExample} is an example of the combined sentences based on the second strategy over the IRI sentence in Ex1.}

\mr{The combined document aims at capturing the correlation between IRIs and words, such as \textit{vc:FOOD-4001 (blonde beer)} and ``nutrient'' in Ex7. 
On the one hand this would benefit the embeddings of the IRIs with the semantics of words.
This is especially useful in some contexts where only IRI vectors are available.
For example, some entities have neither English labels or meaningful IRI name, and only IRI vectors can be used for these entities.
On the other hand, the association with IRIs would incorporate some semantics of the graph structure into the words' embeddings.
Again there are some contexts where only words are analyzed.
One example is when \owlvec is used as an ontology (domain) tailored word embedding model for the classification of external text of this specific domain.
Meanwhile, this may also add noise to the correlation between words (e.g., \textit{vc:hasNutrient} between ``beer'' and ``vitamin'' in Ex9) and negatively impact the words' embeddings.
}
The impact of the combined document and its two strategies is analyzed in our evaluation (cf. Section~\ref{sec:doc}).

\subsection{Embeddings}
\owlvec first merges the structure document, the lexical document and the combined document as one document, 
and then uses this document to train a \textit{Word2Vec} model with 
the skip-gram architecture.
The training is ended when the loss trends to be stable.
The hyper-parameter of the minimum count of words is set to $1$ such that each word or entity (IRI) is encoded as long as it appears in the documents at least once.
Specially, we can pre-train the \textit{Word2Vec} model by a large and general corpus such as a dump of Wikipedia articles.
This brings some prior correlations between words, especially between a word's synonyms and between a word's variants, which enables the downstream machine learning tasks to identify their semantic equality or similarity w.r.t. the corpus.
However, such prior correlations may also be noisy and play a negative role in a domain specific task (cf. the evaluation in Section \ref{sec:pt}).
\mr{Note \textit{Word2Vec} is selected because it is successful, being one of the most widely used word embedding algorithms.
It has already been successfully applied in KG embedding in a combination with random walk; one typical example is RDF2Vec \cite{ristoski2016rdf2vec,ristoski2019rdf2vec}. 
With the adoption of a mature embedding technique, in this study we can focus on extending semantic embedding from a KG to an ontology which expresses a much wider range of semantics, by developing suitable corpus extraction methods.
\owlvec is uncoupled to \textit{Word2Vec}, and is thus compatible with other word embedding or sequence feature learning methods such as the contextual model BERT which has shown its superiority according to some recent studies \cite{miaschi2020contextual}.
We would leave the selection, evaluation or even development of more suitable ad-hoc embedding models in our next study.}

With the trained \mr{word embedding} model,
\owlvec calculates the embedding of each target entity $e$ in $E$.
Its embedding $\bm{e}$ is the concatenation of $V_{iri}(e)$ and $V_{word}(e)$, where $V_{iri}(e)$ is the vector of the IRI of $e$, \mr{and $V_{word}(e)$ is some summarization of the vectors of all the lowercase word tokens of $e$.
In our evaluation we simply adopt the averaging operator for $V_{word}(e)$, which usually works quite well for different data and tasks.
As predictive information of different words' embeddings lie in different dimensions, the averaging operation would not lead to a loss of predictive information, especially when a classifier is further stacked after the embeddings for downstream applications (cf. Section \ref{sec:cs}).
Note some more complicated weighting strategies such as using TF-IDF (term frequency–inverse document frequency) \cite{rajaraman2011data} to calculate the importance of each token can also be considered (cf. \cite{arora2019simple} for more methods).}
As in the case of constructing lexical sentences from IRI sentences, the word tokens of $e$ are extracted from its English label if such a label exists, or from its IRI name otherwise.
Due to the concatenation, the embedding size of $\bm{e}$, i.e., $d$, is twice the original embedding size.
$V_{iri}(e)$ and $V_{word}(e)$ can also be independently used. A comparison of their performance can be found in \ejr{Section~\ref{sec:doc}}. 

\subsection{Case Studies}\label{sec:cs}
We applied \owlvec in ontology completion which first trains a prediction model from known relations (axioms) and then predicts those plausible relations.\footnote{Our ontology completion task is different from ontology reasoning. Our goal is not to infer relations that logically follows from the given input, but to try to discover plausible relations that complement the original ontology. (Most) plausible relations may not be inferred, and our evaluation focuses exactly on those plausible relations that cannot be inferred.}
It includes two tasks: class membership prediction and class subsumption prediction, where the embedding of an entity can be understood as the features automatically learned from its neighbourhood, relevant axioms and lexical information without any supervision. 
\mr{In the remainder of this sub-section we first introduce the prediction details with the membership case and then present the difference of the subsumption case.}

Given a head entity $e_1$ and a tail entity $e_2$, where $e_1$ is an instance and $e_2$ is a class, the membership prediction task aims at training a model to predict the plausibility that $e_1$ is a member of $e_2$ (i.e., $e_2(e_1)$).
The input is the concatenation of the embeddings of $e_1$ and $e_2$, i.e., $\bm{x} = \left[\bm{e_1}, \bm{e_2} \right]$, while the output is a score $y$ in $\left[0, 1\right]$, 
where a higher $y$ indicates a more plausible membership relation.
\mr{For the prediction model, some (non-linear) binary classifiers such as Random Forest (RF) and 
Multi-Layer Perception (MLP) can be adopted (cf. the evaluation of classifiers in Section \ref{sec:classifier}).}

\mr{In training, positive training samples are those declared membership axioms.
They are directly extracted from the ontology.
While negative samples are constructed by corrupting each positive sample.}
Namely, for each positive sample $(e_1, e_2)$, one negative sample $(e_1, e_2^{\prime})$ is generated, where $e_2^{\prime}$ is a random class of the ontology and $e_1$ is not a member of $e_2^{\prime}$ even after entailment reasoning.
In prediction, given a head entity (i.e., the target), a candidate set of classes are selected (e.g., all the classes except for the top class \textit{owl:Thing}, or a subset after filtering via some heuristic rules), each candidate is predicted with a \mr{normalized score by the trained classifier, which indicates the degree of the candidate to be true}, and the candidates are then ranked according to their scores where the top is the most likely class of the instance.
Class subsumption prediction is similar to class membership prediction, except that $e_1$ and $e_2$ are both classes, the goal is to predict whether $e_1$ is subsumed by $e_2$ (i.e., $e_1 \sqsubseteq e_2$), and the head entity $e_1$ itself is excluded from the candidate classes.

\section{Evaluation}\label{sec:evaluation}

\subsection{Experimental Setting}

We evaluated \owlvec on class membership prediction with the HeLis\footnote{HeLis project: \url{https://horus-ai.fbk.eu/helis/}}
ontology~\cite{dragoni2018helis}, and on class subsumption prediction with the 
FoodOn\footnote{FoodOn project: \url{https://foodon.org/}} ontology \cite{dooley2018foodon} and the Gene ontology (GO)\footnote{GO was accessed on August 05, 2020 via \url{http://www.geneontology.org/ontol ogy/}}.
HeLis captures general knowledge about both food and healthy lifestyles, FoodOn captures more detailed knowledge about food, and GO is a major bioinformatics initiative to unify the representation of gene and gene product attributes.
\mr{Their DL expressivities are $\mathcal{ALCHIQ(D)}$, $\mathcal{SRIQ}$ and $\mathcal{SRI}$ respectively.
Some statistics of the two ontologies are shown in Table~\ref{res:statistics}.}
Due to different knowledge representations, HeLis has a large number of membership axioms but a very small number of subsumption axioms, while FoodOn and GO have only subsumptions axioms.
This is the reason why we evaluated membership prediction on HeLis, but subsumption prediction on FoodOn and Go.
Data and codes are available at  \url{https://github.com/KRR-Oxford/OWL2Vec-Star}.

\begin{table}[t]
\scriptsize{
\centering
\renewcommand{\arraystretch}{1.3}
\begin{tabular}[t]{p{0.85cm}<{\centering}|p{0.9cm}<{\centering}|p{0.9cm}<{\centering}|p{1.05cm}<{\centering}|p{0.85cm}<{\centering}|p{0.9cm}<{\centering}|p{0.95cm}<{\centering}|p{1.35cm}<{\centering}|p{0.96cm}<{\centering}|p{1cm}<{\centering}}\hline
&\tabincell{Instances\\\#} & \tabincell{Classes \\ \#} & \tabincell{Axioms \\ \#} & \tabincell{Memb. \\ \#} & \tabincell{Subs. \\ \#} & \tabincell{ Label \\ Avg Size} & \tabincell{Entities with \\ No Label} & \tabincell{Anno. \\ Avg Size} & \tabincell{Anno. \# \\ per Entity}\\ \hline
HeLis &  $20,318$ & $277$ & $172,213$ & $20,318$ & $261$ &$1.71$ &$0.05\%$ & $0.13$ & $0.14$ \\ \hline
FoodOn &  $359$ & $28,182$  & $241,581$ & $0$ & $29,778$ &$2.92$ &$0.32\%$ &$6.70$ & $3.99$ \\ \hline
GO &  $0$ & $44,244$  & $513,306$ & $0$ & $72,601$ &$4.26$ & $0.15\%$ & $4.26$ &$9.09$ \\ \hline
\end{tabular}
\vspace{-0.1cm}
\caption{\footnotesize
\mr{Statistics of the HeLis ontology, the FoodOn ontology and the GO ontology. Memb. and Subs. denote the membership axiom and the subsumption axiom respectively\protect\footnotemark. \# denotes the count. Size refers to the number of word tokens. Label refers to the English label by \textit{rdfs:label} of the instance or the class, or the meaningful IRI name if their is no such an English label. Anno. refers to the literals by annotation properties (excluding \textit{rdfs:label}). Entity here refers to the instance and the class. }}\label{res:statistics}
}
\end{table}
\footnotetext{Membership and subsumption in Table \ref{res:statistics} denote the declared membership and subsumption axioms with named classes alone, i.e., those involving composed classes and those inferred are not counted.}


The experiment on membership and subsumption prediction follows the following setting: all the explicitly declared class membership axioms (or class subsumption axioms) are randomly divided into three sets for training ($70\%$), validation ($10\%$) and testing ($20\%$), respectively.
For each axiom in the validation/testing set, the head entity (i.e., an instance for membership prediction and a class for subsumption prediction) is the target whose class is to be predicted from all the candidates and compared against the tail entity (as the ground truth class) in evaluation.
All the candidates are ranked according to the predicted score which indicates the likelihood of being the head entity's class.
We calculate the following widely adopted metrics: Hits@$1$, Hits@$5$, Hits@$10$ and MRR (Mean Reciprocal Rank).
The first three measure the recall of the ground truths within the top $1$, $5$ and $10$ ranking positions, respectively,
while the fourth averages the reciprocals of the ranking positions of the ground truths.
The higher the metrics, the better the performance.

The performance of \owlvec is reported with the following settings.
For the \mr{embedding model, its} dimension is set to $100$ if no pre-training is adopted, and otherwise set to be consistent with the pre-trained model; 
the window size is set to $5$; the minimum count of words is set to $1$; the iteration number of training is set to $10$, which is based on the observation of the loss.
\mr{The \textit{Word2Vec} pre-training (with a dimension of $200$) uses the latest English Wikipedia article dump\footnote{\url{https://dumps.wikimedia.org/enwiki/}}, as in many other \textit{Word2Vec} relevant studies such as \cite{chen2019canonicalizing}.
Other corpus or pre-trained models can also be used, and we can further select a corpus that is specific to the domain of the ontology.
Such extensive evaluation will be considered in the future work.
Random Forest (RF) is adopted as the basic binary classifier and the WL sub-graph kernel is enabled in the random walk if they are not specified.}
Other hyper-parameters such the walking depth and the transformation from OWL ontology to RDF graph, as well as the hyper-parameters of the baselines are adjusted through the validation set --- the setting that leads to the highest MRR is adopted.

The evaluation is organized as follows.
We first compare \owlvec with the baselines, 
\mr{then analyze the impact of different settings including the documents, the IRI and word embeddings, the settings for generating the structure document (walking type, walking depth and the transformation from OWL ontology to RDF graph), the usage of reasoning and pre-training. 
We next analyze the effectiveness of \owlvec towards different classifiers including RF, MLP, Logistic Regression (LR) and Support Vector Classifier (SVC), all of which are implemented by scikit-learn \cite{pedregosa2011scikit},}
and finally analyze the embeddings via visualization and comparing the Euclidean distances.
The selected embedding baselines include \textit{(i)} four well-known knowledge graph embedding methods, i.e., RDF2Vec, TransE, TransR and DistMult, 
\textit{(ii)} four state-of-the-art ontology embedding methods, i.e., Onto2Vec, OPA2Vec, EL Embedding and Quantum Embedding,\footnote{For EL (resp. Quantum) Embedding, HeLis and FoodOn are first transformed into  DL $\mathcal{EL}^{++}$ (resp. DL $\mathcal{ALC}$) by removing logical axioms outside the supported expressivity. }
\textit{(iii)} the original OWL2Vec which is equivalent to \owlvec using the IRI embedding, structure document and ontology projection rules,
\textit{and (iv)} the pre-trained \textit{Word2Vec} model.
The embeddings of these baselines are applied to the two tasks in the same way as \owlvec, \mr{with the Random Forest classifier.}
Note that RDF2Vec, TransE, TransR and DistMult are trained with the RDF graph $\mathcal{G}$ transformed from the original ontology using OWL to RDF Graph Mapping without entailment reasoning, 
while the pre-trained \textit{Word2Vec} calculates the average word vector of an entity according to its label (or its IRI name if the label does not exist) as in \owlvec.

\mr{
We realized the surface form of the textual information may play a role in our prediction tasks, as the ontology engineer may follow some form in naming a class's the members or subclasses.
For example, the instance ``Blonde Beer'' is named by adding a prefix to the name of its class ``Beer''.
Thus we further compared our \owlvec plus RF solution to the supervised Transformer classifier \cite{vaswani2017attention} which embed the head and tail entities' contextual text.
The Transformer classifier has two versions: \textit{label} which considers the English labels and IRI names of the two entities, and \textit{all text} which considers all of the two entities' textual labels and annotations.
The label, IRI name and textual annotation are pre-processed in the same was as in learning \owlvec, and they are orderly concatenated into one sequence as the input of the Transformer.

For RDF2Vec we use the implementation of pyRDF2Vec\footnote{\url{https://github.com/IBCNServices/pyRDF2Vec}}; for TransE, TransR and DistMult we use the implementation of OpenKE\footnote{\url{https://github.com/thunlp/OpenKE}}; for Onto2Vec and OPA2Vec we implement them as special cases of \owlvec; for EL Embedding and Quantum Embedding we use the codes attached in their original papers \cite{holter2019embedding} and \cite{garg2019quantum}, respectively. 
The Transformer classifier is implemented by Tensorflow\footnote{\url{https://www.tensorflow.org/}} with one token and position embedding layer, and one Transformer block that contains two attention heads.
All the results are generated locally with repetitions. 
}


\subsection{Comparison with Baselines} 
\label{sec:eval-baselines}
\mr{Table \ref{res:overall} reports the performance of \owlvec and the baselines with their optimum settings.
The performance of \owlvec with different settings can be found in Section \ref{sec:ea}.
In Table \ref{res:overall} we can observe that \owlvec outperforms all the baselines.
Note all these comparisons have statistical significance with the \textit{p}-value being $\ll 0.05$ in the two-tailed test.}
Among all these ontology embedding and KG embedding baselines which directly calculate the IRI's vector without considering the word vector, OPA2Vec achieves the best performance on FoodOn and GO for subsumption prediction; while the KG embedding method RDF2Vec performs the best on HeLis for class membership prediction.
In contrast, the two logic embedding methods Quantum Embedding and EL Embedding, and TransE perform poorly on all the three ontologies.
Our preliminary work OWL2Vec achieves promising results on HeLis (close to RDF2Vec) and FoodOn (close to OPA2Vec), but performs poorly on GO.
\owlvec outperforms both KG embedding methods and  ontology embedding methods; for example, \mr{consider the Hits@$1$ of \owlvec, it is $325.6\%$ higher than RDF2Vec on HeLis, $146.6\%$ higher than OPA2Vec on FoodOn, and $126.7\%$ higher than OPA2Vec on GO.}

Meanwhile, \owlvec outperforms the pre-trained \textit{Word2Vec}, with $6.0\%$, $56.6\%$ and $38.2\%$ higher MRR on HeLis, FoodOn and GO, respectively.
It is interesting to see that  the pre-trained \textit{Word2Vec} using entity labels or IRI names achieves good performance, outperforming those ontology and KG embedding baselines such as RDF2Vec and OPA2Vec.
It means that the textual information plays a very important role in embedding real world ontologies, especially for membership prediction and subsumption prediction as the names of the instances and classes with a membership or subsumption relationship often use some common words or \mr{words with relevant meanings (e.g., synonyms or word variants).
This observation on the importance of the textual information is consistent with our following ablation study on the usage of word embedding, IRI embedding and both (cf. $V_{iri}$, $V_{word}$ and $V_{iri,word}$ in Table \ref{res:setting}).}
A key difference between \owlvec and \textit{Word2Vec} is that the word embedding model of \owlvec is~trained by an ontology tailored corpus underpinned by its graph structure and logical axioms.

\mr{In Table \ref{res:overall}, we can also observe that both Transformer classifiers (i.e., \textit{label} and \textit{all text}) perform worse than the RF classifier using the pre-trained \textit{Word2Vec} or \owlvec embeddings.
Specially, on HeLis, they are effective with some promising results which are better than the KG and ontology embedding baselines; while on FoodOn and GO, they are very ineffective, with much worse performance than all the other methods. 
This means that the surface form of the entities' textual contexts (token sequences), with feature learning by Transformer, brings very little predictive information on FoodOn and GO and partial predictive information on HeLis.
This result is consistent with our manual observation on the ontologies' naming mechanisms on the instances and subclasses, 
and it in turn verifies that the semantics from the large Wikipedia corpus encoded in the \textit{Word2Vec} embeddings, and the semantics of the ontology graph and logical constructors encoded in the \owlvec embeddings play a very important role in these prediction tasks.}

Note that the performance of membership prediction with HeLis is much higher than that of the subsumption prediction with FoodOn and Go.
This is because the former has much less candidate classes (cf. Table \ref{res:statistics}) and is thus less challenging.
\mr{Meanwhile the entity name and label's surface form (i.e., the class member's naming mechanism) of HeLis makes additional contribution to its membership prediction, as analyzed above.}


\begin{table}[t]
\centering
\scriptsize{
\renewcommand{\arraystretch}{1.3}
\begin{subtable}[h]{\textwidth}
\centering
\begin{tabular}[t]{c||p{1.1cm}<{\centering}|p{0.9cm}<{\centering}|p{0.90cm}<{\centering}|p{0.90cm}<{\centering}}\hline
&\multicolumn{4}{c}{HeLis} \\\cline{2-5}
Method& MRR & Hits@$1$ & Hits@$5$ & Hits@$10$ \\ \hline
\mr{Transformer (\textit{label})} & $0.657$ & $0.515$ & $0.824$  &$0.897$  \\ \hline
\mr{Transformer (\textit{all text})} & $0.599$  & $0.390$  &$0.870$  &$0.912$  \\ \hline\hline
RDF2Vec & $0.345$ & $0.219$ &$0.460$  &$0.655$  \\ \hline
TransE & $0.181$&  $0.09$ & $0.232$ & $0.355$ \\ \hline
TransR & $0.298$& $0.184$ &$0.391$  &$0.559$  \\ \hline
DistMult &$0.253$ & $0.166$ & $0.304$ &$0.437$   \\ \hline\hline
Quantum Embeding &$0.159$ &$0.132$  &$0.163$  &$0.190$ \\ \hline
Onto2Vec &$0.211$ & $0.108$ & $0.268$ &$0.397$  \\ \hline
OPA2Vec &$0.237$ &$0.146$  & $0.286$ &$0.408$   \\ \hline
OWL2Vec &$0.335$ &$0.215$  &$0.397$  & $0.601$  \\ \hline \hline
Pre-trained \textit{Word2Vec} &$0.899$&$0.877$& $0.923$& $0.933$  \\ \hline \hline
\owlvec & $\bm{0.953}$ & $\bm{0.932}$ & $\bm{0.978}$ & $\bm{0.987}$  \\ \hline
\end{tabular}
\vspace{-0.1cm}
\caption{Membership Prediction}
\end{subtable}
\hfill
\vspace{0.15cm}
\begin{subtable}[h]{\textwidth}
\centering
\begin{tabular}[t]{c||p{0.9cm}<{\centering}|p{0.90cm}<{\centering}|p{0.90cm}<{\centering}|p{0.90cm}<{\centering}||p{0.90cm}<{\centering}|p{0.90cm}<{\centering}|p{0.90cm}<{\centering}|p{0.90cm}<{\centering}}\hline
&\multicolumn{4}{c||}{FoodOn} &  \multicolumn{4}{c}{GO} \\\cline{2-9}
Method& MRR & Hits@$1$ & Hits@$5$ & Hits@$10$ & MRR & Hits@$1$ & Hits@$5$& Hits@$10$\\ \hline
\mr{Transformer (\textit{label})} &$0.016$  & $0.005$ & $0.027$ & $0.046$ &$0.009$ &$0.001$  &$0.009$ & $0.018$ \\ \hline
\mr{Transformer (\textit{all text})} & $0.022$   & $0.011$&$0.032$ &$0.050$ &$0.014$ &$0.008$ &$0.017$ &$0.019$  \\ \hline\hline
RDF2Vec &  $0.078$& $0.053$& $0.097$&$0.119$ & $0.043$& $0.017$& $0.057$&$0.087$  \\ \hline
TransE & $0.029$ & $0.011$&$0.044$ & $0.065$&$0.015$ & $0.005$&$0.018$ & $0.030$ \\ \hline
TransR & $0.072$ & $0.044$&$0.093$ &$0.130$ &$0.048$ & $0.016$&$0.076$ &$0.113$  \\ \hline
DistMult &$0.076$ & $0.045$& $0.099$&$0.134$ &$0.046$ & $0.018$& $0.68$&$0.097$   \\ \hline\hline
EL Embeding & $0.040$ & $0.014$&$0.067$ &$0.099$ & $0.018$ & $0.005$&$0.021$ &$0.036$  \\ \hline
Onto2Vec &$0.034$ & $0.014$& $0.047$& $0.064$ &$0.024$ & $0.008$& $0.031$& $0.053$  \\ \hline
OPA2Vec &$0.093$ &$0.058$ &$0.117$ &$0.156$ &$0.075$ &$0.032$ &$0.106$ &$0.157$   \\ \hline
OWL2Vec &$0.091$ &$0.052$  &$0.121$  &$0.152$ &$0.031$ &$0.012$ &$0.040$ &$0.067$  \\ \hline \hline
Pre-trained \textit{Word2Vec} &$0.136$& $0.089$& $0.175$& $0.227$ & $0.123$& $0.055$& $0.177$& $0.260$   \\ \hline \hline
\owlvec & $\bm{0.213}$ &$\bm{0.143}$ &$\bm{0.287}$ &$\bm{0.357}$&$\bm{0.170}$ &$\bm{0.076}$ &$\bm{0.258}$ &$\bm{0.376}$  \\ \hline
\end{tabular}
\vspace{-0.1cm}
\caption{Subsumption Prediction}
\end{subtable}
\caption{\footnotesize
Overall results of \owlvec and the baselines.
}\label{res:overall}
}
\end{table}

\subsection{Analysis of \owlvec 
Settings}\label{sec:ea}
\subsubsection{Lexical Information}\label{sec:doc}
According to Table \ref{res:setting} we can find that the lexical document $D_{l}$ leads to a significant improvement of performance when it is merged with the structure document $D_{s}$ (\ie  $D_{s,l}$).
Consider MRR, $D_{s,l}$ outperforms $D_{s}$ by $26.9\%$ on HeLis, by $18.8\%$ on FoodOn and by $22.1\%$ on GO when the IRI embedding ($V_{iri}$) is used,
and by $169.7\%$, $31.8\%$ and $44.2\%$ respectively when both IRI embedding and word embedding ($V_{iri,word}$) are used.

Unlike the lexical document, the combined documents ($D_{s,l,rc}$ and $D_{s,l,tc}$), which also rely on the lexical information of the ontology, lead to a limited positive impact.
For class membership prediction, the best performance of $D_{s,l,rc}$ \mr{(MRR: $0.951$)} and the best performance of $D_{s,l,tc}$ \mr{(MRR: $0.953$)} are both very close to the best performance as $D_{s,l}$ \mr{(MRR: $0.952$)}, 
while for class subsumption prediction on FoodOn and GO, they are both worse than the best performance of $D_{s,l}$.
We can also find that the combined document has a negative impact when the word embedding ($V_{word}$) is used alone on FoodOn and GO.
\mr{On FoodOn, $D_{rc}$ ($D_{tc}$ resp.) reduces the MRR from $0.213$ to $0.196$ ($0.194$ resp.), while on GO, $D_{rc}$ ($D_{tc}$ resp.) reduces the MRR from $0.170$ to $0.155$ ($0.150$ resp.).}
This is because the combined sentences build the correlation between words and IRIs, which benefits the IRI embeddings, but brings noise to the correlation between words and harms the word embeddings.

\begin{table}[t]
\scriptsize{
\renewcommand{\arraystretch}{1.3}
\begin{subtable}[h]{\textwidth}
\centering
\begin{tabular}[t]{p{2.6cm}<{\centering}||p{1cm}<{\centering}|p{1cm}<{\centering}|p{1cm}<{\centering}|p{1cm}<{\centering}}\hline
&\multicolumn{4}{c}{HeLis} \\\cline{2-5}
Setting& MRR & Hits@$1$ & Hits@$5$ & Hits@$10$ \\ \hline
$D_{s}$ + $V_{iri}$ & $0.353$& $0.226$& $0.470$&$0.668$  \\ \hline\hline
$D_{s,l}$ + $V_{iri}$ &$0.448$ &$0.295$ &$0.623$ &$0.814$  \\ \hline
$D_{s,l}$ + $V_{word}$ &$0.938$ &$0.920$ &$0.961$ &$0.974$  \\ \hline
$D_{s,l}$ + $V_{iri,word}$ & $0.952$ &$\bm{0.934}$ &$0.974$ &$0.984$ \\ \hline\hline
$D_{s,l,rc}$ + $V_{iri}$ & $0.446$ &$0.299$ & $0.618$& $0.799$\\ \hline
$D_{s,l,rc}$ + $V_{word}$ & $0.945$&$0.926$ &$0.970$ &$0.979$ \\ \hline
$D_{s,l,rc}$ + $V_{iri,word}$ &$0.951$ &$0.932$ & $\bm{0.975}$&$\bm{0.987}$  \\ \hline\hline
$D_{s,l,tc}$ + $V_{iri}$ &$0.505$ & $0.355$& $0.695$&$0.854$ \\ \hline
$D_{s,l,tc}$ + $V_{word}$ &$0.943$ &$0.923$ &$0.969$ & $0.976$ \\ \hline
$D_{s,l,tc}$ + $V_{iri,word}$ & $\bm{0.953}$& $0.932$&$\bm{0.975}$ &$\bm{0.987}$  \\ \hline
\end{tabular}
\vspace{-0.1cm}
\caption{Membership Prediction}
\hfill
\vspace{0.1cm}
\end{subtable}
\begin{subtable}[h]{\textwidth}
\centering
\begin{tabular}[t]{p{2.3cm}<{\centering}||p{0.7cm}<{\centering}|p{0.7cm}<{\centering}|p{0.7cm}<{\centering}|p{0.85cm}<{\centering}||p{0.7cm}<{\centering}|p{0.7cm}<{\centering}|p{0.7cm}<{\centering}|p{0.8cm}<{\centering}}\hline
&\multicolumn{4}{c||}{FoodOn} &  \multicolumn{4}{c}{GO} \\\cline{2-9}
Setting& MRR & Hits@$1$ & Hits@$5$ & Hits@$10$ & MRR & Hits@$1$ & Hits@$5$& Hits@$10$\\ \hline
$D_{s}$ + $V_{iri}$ & $0.154$ &$0.102$&$0.210$ & $0.150$ &$0.095$ &$0.044$&$0.144$ & $0.195$ \\ \hline\hline
$D_{s,l}$ + $V_{iri}$ &$0.183$ &$0.120$&$0.249$ & $0.305$ &$0.116$ &$0.048$&$0.185$ & $0.252$ \\ \hline
$D_{s,l}$ + $V_{word}$ &$\bm{0.213}$ &$\bm{0.143}$&$\bm{0.287}$ & $\bm{0.357}$ &$\bm{0.170}$ &$\bm{0.076}$&$\bm{0.258}$ & $\bm{0.376}$ \\ \hline
$D_{s,l}$ + $V_{iri,word}$ & $0.203$ &$0.133$&$0.273$ & $0.338$ &$0.137$ &$0.068$&$0.204$ & $0.319$\\ \hline\hline
$D_{s,l,rc}$ + $V_{iri}$ & $0.188$ &$0.125$&$0.249$ & $0.310$& $0.115$ &$0.050$&$0.164$ & $0.249$ \\ \hline
$D_{s,l,rc}$ + $V_{word}$ & $0.196$ &$0.127$& $0.246$& $0.329$ & $0.155$ &$0.066$& $0.237$& $0.348$ \\ \hline
$D_{s,l,rc}$ + $V_{iri,word}$ &$0.201$&$0.134$&$0.270$ & $0.330$ & $0.139$&$0.060$&$0.206$ & $0.297$ \\ \hline\hline
$D_{s,l,tc}$ + $V_{iri}$ &$0.172$&$0.107$&$0.234$ &$0.297$ & $0.101$&$0.046$&$0.154$ &$0.209$  \\ \hline
$D_{s,l,tc}$ + $V_{word}$ &$0.194$ &$0.130$& $0.254$&$0.314$ &$0.150$ &$0.061$& $0.230$&$0.343$  \\ \hline
$D_{s,l,tc}$ + $V_{iri,word}$ & $0.202$ &$0.127$&$0.278$ &$0.349$ &$0.139$ &$0.061$&$0.204$ &$0.300$ \\ \hline
\end{tabular}
\vspace{-0.1cm}
\caption{Subsumption Prediction}
\end{subtable}
\caption{\footnotesize
The results of \owlvec under different document ($D$) and embedding ($V$) settings.
Subscripts: $s$ (resp. $l$) denotes the structure (resp. lexical) document;
$rc$ (resp. $tc$) denotes the combined document with the random (resp. traversal) strategy; $iri$ (resp. $word$) denotes the IRI (resp. word) embedding.
}\label{res:setting}
}
\end{table}

Besides the lexical document, the word embedding ($V_{word}$) which also benefits from the utilization of the lexical information of the ontology shows a very strong positive impact.
On the one hand, \ejr{as discussed in Section \ref{sec:eval-baselines}},
the two methods that use the word embedding, i.e., \owlvec and the pre-trained \textit{Word2Vec}, both dramatically outperform the remaining methods.
On the other hand, as shown in Table \ref{res:setting}, the best performance on HeLis comes from $V_{iri,word}$, while the best performance on FoodOn and GO comes from $V_{word}$.
The outperformance of $V_{iri,word}$ and $V_{word}$ over $V_{iri}$ is quite significant; for example, \mr{when the lexical and structure documents ($D_{s,l}$) are used, the Hits@$1$ of $V_{iri,word}$ is $0.934$, $0.133$ and $0.068$ on HeLis, FoodOn and GO, respectively,  while the corresponding Hits@$1$ of $V_{iri}$ is $0.295$, $0.120$ and $0.048$, respectively.}

Regarding the IRI embedding, on the one hand it can alone outperform the baseline embeddings in Table \ref{res:overall} except for the pre-trained \textit{Word2Vec}.
On the other hand, the impact of the IRI embedding when it is concatenated with the word embedding varies from task to task.
It has a positive impact on class membership prediction with HeLis;
for example, when trained by the structure document and lexical document ($D_{s,l}$), the MRR of $V_{iri,word}$ is $1.5\%$ higher than $V_{word}$.
However, on class subsumption prediction with FoodOn and GO, the IRI embedding shows a negative impact, \mr{i.e., $V_{iri, word}$ is often close to or worse than $V_{word}$.
This may be due to the fact that the lexical semantics plays a dominant role in these prediction tasks, and the word sentences from which the word embeddings are learned have already used the structure sentences as the backbone. 
Meanwhile, we simply concatenate the two vectors as the input of a basic classifier without any mechanisms to better integrate the two embeddings (inputs).
}

\subsubsection{Graph Structure} \label{sec:gs}
\begin{figure}[t]
\begin{center}
\begin{subfigure}[b]{\textwidth}
\centering
\includegraphics[width=0.86\textwidth]{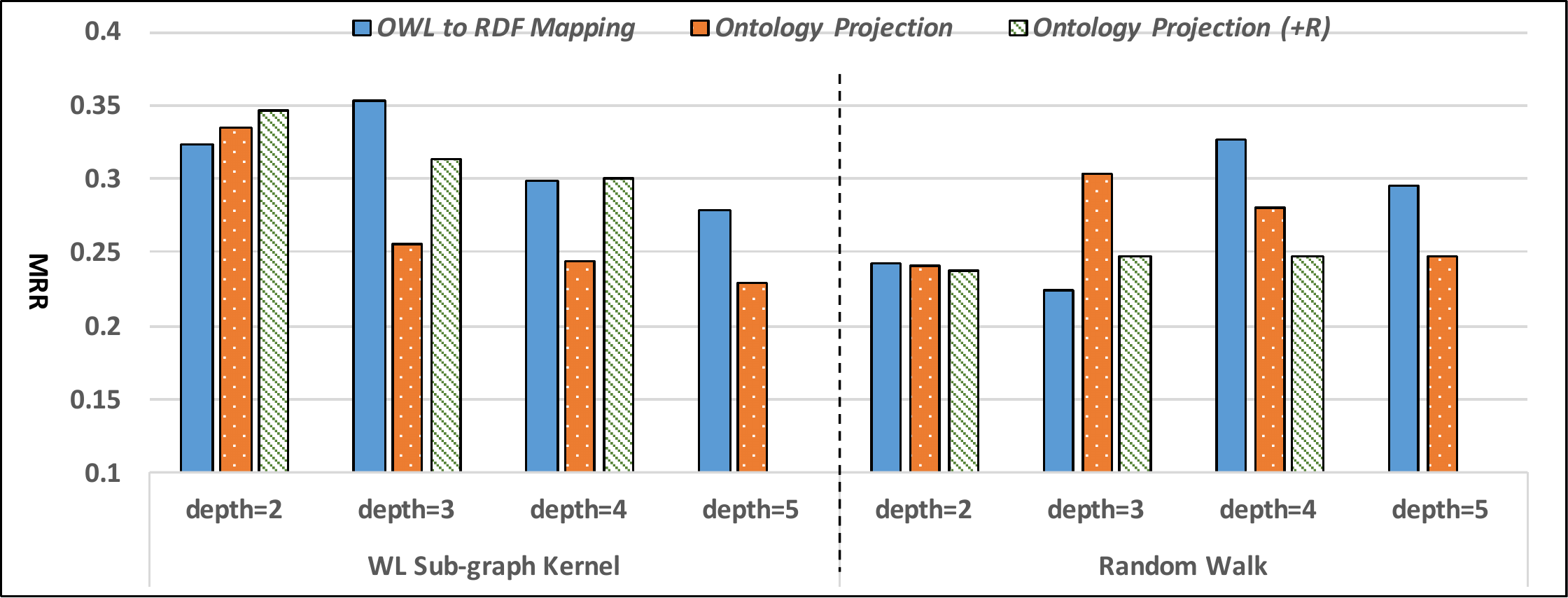}
\vspace{-0.1cm}
\caption{Membership Prediction (HeLis)}
\end{subfigure}
\hfill
\vspace{0.03cm}
\begin{subfigure}[b]{\textwidth}
\centering
\includegraphics[width=0.86\textwidth]{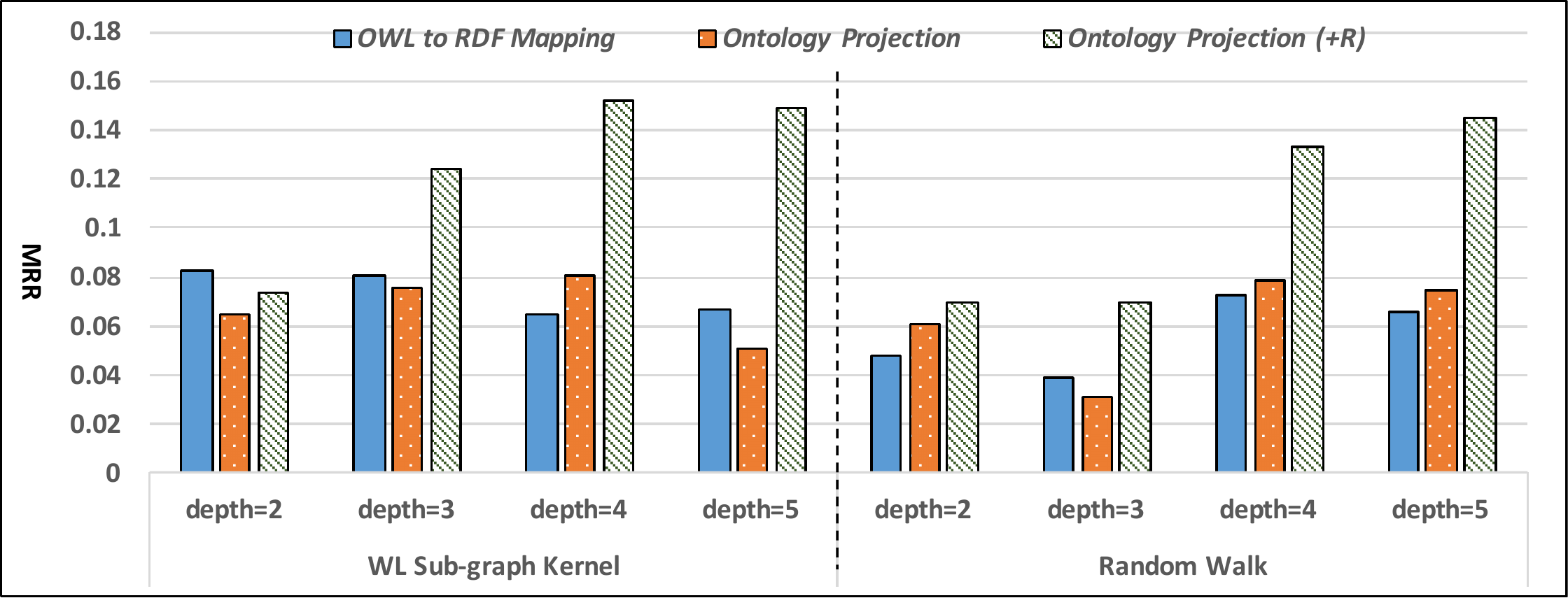}
\vspace{-0.1cm}
\caption{Subsumption Prediction (FoodOn)}
\end{subfigure}
\hfill
\vspace{0.03cm}
\begin{subfigure}[b]{\textwidth}
\centering
\includegraphics[width=0.86\textwidth]{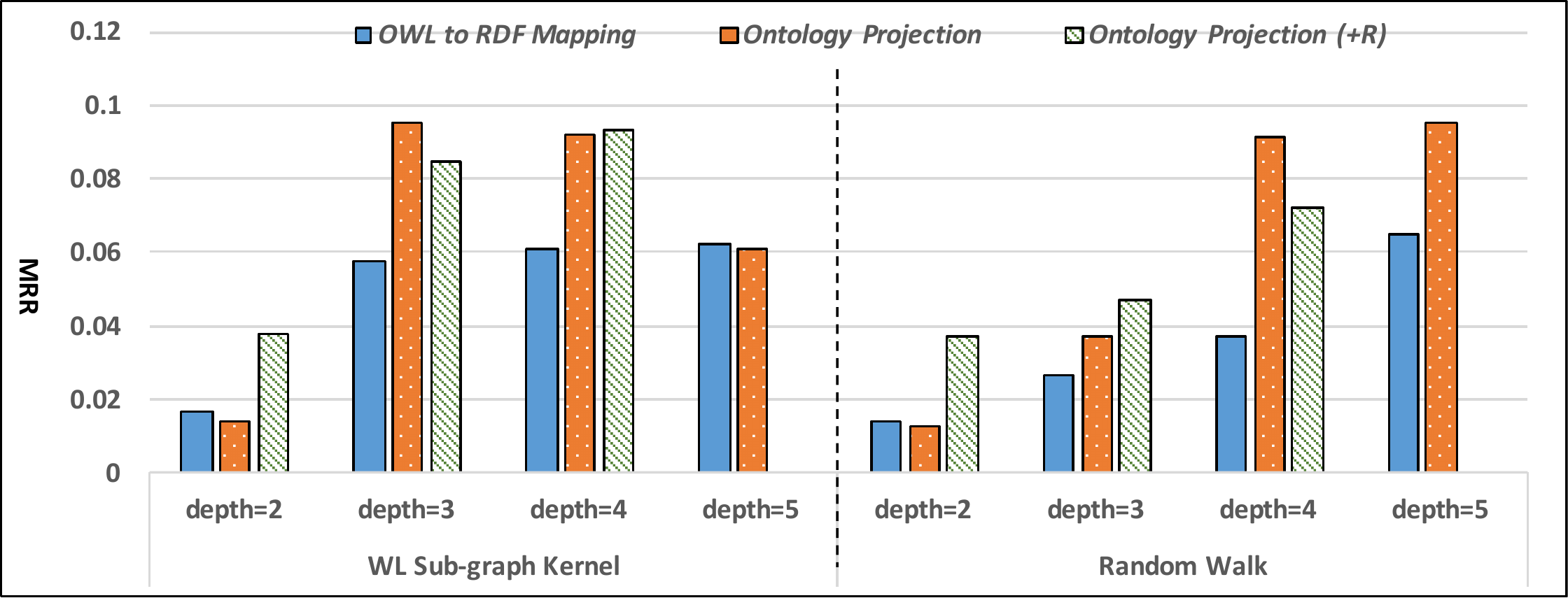}
\vspace{-0.1cm}
\caption{Subsumption Prediction (GO)}
\end{subfigure}
\vspace{-0.2cm}
\caption{\footnotesize Comparison of structure documents by different graph structure exploration settings, with the MRR results of \owlvec($D_{s}$ + $V_{iri}$) reported.}
\label{fig:graph}
\end{center}
\end{figure}

Figure \ref{fig:graph} shows the performance of the IRI embedding of \owlvec when it is trained using structure documents extracted under different ontology graph structure exploration settings.
We first compare the two solutions that generate the RDF graph $\mathcal{G}$: 
\mr{\textit{(i)} the \owl to RDF Graph Mapping defined by W3C, which may lead to redundant blank nodes and longer paths some complex axioms but keeps the complete semantics, 
\textit{(ii)} the ontology projection rules which lead to a more compact graph but approximate most complex axioms, i.e., some logical relationships are missed in the projected RDF graph, and \textit{(iii)} the ontology projection rules with inverse triples for the membership and subsumption axioms. Please see Section \ref{sec:otog} for more details.
Note that some results of the solution \textit{(iii)} with the depth of $5$ are missing as the walking does not stop after hours.
On HeLis, the solution \textit{(i)} has higher MRR when the WL sub-graph kernel is enabled and the depth is set to $\geq 3$, or when the original random walk is adopted and the depth is set to $\geq 4$.
Its best MRR value (i.e., $0.353$) is higher than the other solutions (i.e., $0.335$ and $0.346$).
On FoodOn, the solution \textit{(iii)} has much higher MRR when the walking depth is $\geq 4$ (as high as $0.152$ with the WL sub-graph kernel enabled) than the best of the solution \textit{(i)} and \textit{(ii)} ($0.083$ and $0.081$ respectively). 
On GO, the solution \textit{(ii)} performs well when the walking depth is set to $3$ or $4$, and the WL sub-graph kernel is enabled, or when the walking depth is $\geq 4$ with the pure random walk.
It also leads to the best MRR, i.e., $0.095$.
Therefore, the performance of these three ontology to RDF graph transformation methods varies from ontology to ontology; in \owlvec, the OWL to RDF Graph Mapping is adopted on HeLis, the projection rules with inverse are adopted on FoodOn and the projection rules are adopted on GO.}

With Figure \ref{fig:graph} we can also compare different \mr{walking strategies and walking depths} used in extracting IRI sentences from the RDF graph $\mathcal{G}$.
We get two observations.
First, the walking depth is important for both random walk and random walk with the WL sub-graph kernel.
In general, to achieve the best performance, the latter needs a smaller walking depth.
Consider the OWL to RDF Mapping, the optimal walking depth  is $3$ on HeLis and $2$ on FoodOn for random walk with the WL sub-graph kernel, but is $4$ for raw random walk.
\mr{Consider the ontology GO with the projection rules, the best performance for random walk with the WL sub-graph kernel often lies in the depth of $3$, while the best performance for raw random walk lies in the depth of $5$.
Second, the top MRR with the WL sub-graph kernel enabled is higher than the top MRR with raw random walk on HeLis and FoodOn, and is the same on GO.
Both observations are as expected because enabling the WL sub-graph kernel incorporates the structure information of the sub-graphs of partial entities of a random walk.}

\subsubsection{Logical Constructors}\label{sec:lc}
On the one hand, the performance of the baselines in Table \ref{res:overall} which adopt the logical structure alone, including EL Embedding, Quantum Embedding and Onto2Vec, is relatively poor in comparison with the other methods.
On the other hand, the logical structure has a positive impact when it works together with the graph structure in \owlvec.
\mr{On HeLis, in comparison with RDF2Vec, the difference of \owlvec with the setting of the structure document and the IRI embedding (i.e., $D_s$ + $V_{iri}$) is that it additionally uses sentences from Manchester Syntax axioms; while on FoodOn and GO, the difference also includes that \owlvec ($D_s$ + $V_{iri}$) uses the projection rules (with inverse) while RDF2Vec uses OWL to RDF Graph Mapping.
By comparing the results of RDF2Vec in Table \ref{res:overall} and the results of \owlvec ($D_s$ + $V_{iri}$) in Table \ref{res:setting}, we can find the impact of adding these axiom sentences is positive on HeLis; the latter (i.e., using axiom sentences) has $2.3\%$ higher MRR and $3.2\%$ higher Hits@1.
On FoodOn and GO, the outperformance of \owlvec ($D_s$ + $V_{iri}$) over RDF2Vec is very significant, with $97.4\%$ higher MRR and $92.5\%$ higher Hits@1 on FoodOn, and
 $120.9\%$ higher MRR and $164.7\%$ higher Hits@1 on GO.
This outperformance is partially due to the usage of the Manchester Syntax axioms, and is partially due to the projection rules (with inverse).
}

We also analyzed the impact of using reasoning (provided by \owl 2 reasoner HermiT) before the ontology is transformed into an RDF graph.
\mr{The results on Onto2Vec, OPA2Vec, \owlvec ($D_s$ + $V_{iri}$) and \owlvec ($D_{s,l}$ + $V_{word}$) are shown in Table \ref{res:inference}.}
We can see that reasoning has a limited impact in the conducted experiments; the MRR results \textit{with} and \textit{without} reasoning are quite close \mr{for OPA2Vec and \owlvec with both settings of $D_{s}$ + $V_{iri}$ and $D_{s,l}$ + $V_{word}$.
Note that \owlvec in Table \ref{res:inference} uses W3C OWL to RDF Graph Mapping on HeLis, projection rules with inverse on FoodOn and projection rules on GO.
The impact of reasoning is limited with all the three transformation approaches.
The impact of reasoning for Onto2Vec is more significant especially on FoodOn and GO.
That may be because it uses the axiom sentences that are more likely to be impacted by entailment reasoning.
OPA2Vec uses the sentences of literal annotations which are much less impacted by entailment reasoning.
\owlvec uses multiple kinds of sentences and are thus more robust.
Meanwhile, the impact of reasoning on Onto2Vec varies from ontology to ontology; for example, it is positive for FoodOn but negative for GO.}

\begin{table}[t]
\scriptsize{
\centering
\renewcommand{\arraystretch}{1.3}
\begin{tabular}[t]{p{1.1cm}<{\centering}|p{1.1cm}<{\centering}||p{1.2cm}<{\centering}|p{1.2cm}<{\centering}|p{2.8cm}<{\centering}|p{3cm}<{\centering}}\hline
&Setting& Onto2Vec & OPA2Vec &\owlvec ($D_s$ + $V_{iri}$) & \owlvec ($D_{s,l}$ + $V_{word}$)\\ \hline
\multirow{2}{*}{\tabincell{HeLis}} &\textit{with} & $0.211$& $0.237$ &$0.340$ &$0.938$ \\ \cline{2-6}
&\textit{without} &$0.221$ & $0.226$ &$0.353$ &$0.935$  \\ \hline
\multirow{2}{*}{\tabincell{FoodOn}} & \textit{with} &$0.034$ &$0.087$ &$0.152$ &$0.188$ \\ \cline{2-6}
& \textit{without} &$0.019$ &$0.088$ &$0.154$ &$0.205$ \\ \hline
\multirow{2}{*}{\tabincell{GO}} & \textit{with} &$0.024$ &$0.075$ &$0.092$ &$0.167$ \\ \cline{2-6}
& \textit{without} &$0.034$ &$0.087$ &$0.095$ &$0.170$ \\ \hline
\end{tabular}
\vspace{-0.2cm}
\caption{\footnotesize
\mr{Performance (MRR) \textit{with} and \textit{without} entailment reasoning. Regarding \owlvec, the ontology is transformed into an RDF graph with the approach that achieves the best performance, i.e., W3C OWL to RDF Graph Mapping on HeLis, projection rules (+R) for FoodOn, and projection rules for GO.}
}\label{res:inference}
}
\end{table}

\subsubsection{Pre-training}\label{sec:pt}
\mr{
With the setting of $D_{s,l}$ + $V_{word}$, the MRR of \owlvec decreases to $0.933$, $0.136$ and $0.147$ on HeLis, FoodOn and GO respectively, while its Hits@$1$ decreases to $0.913$, $0.091$ and $0.069$ respectively.
On the one hand, using a pre-trained \textit{Word2Vec} \mr{does not increase} but decreases the performance of \owlvec (cf. Table \ref{res:setting} for the corresponding results without pre-training).
That may be because the pre-trained \textit{Word2Vec} is short of prior correlations involving entity IRIs, and its usage also leads to less compact embeddings with their dimension increases from $100$ to $200$.
On the other hand, \owlvec with pre-training still outperforms the original \textit{Word2Vec} whose results are shown in Table \ref{res:overall}.
This in turn verifies that the embeddings learned from}
\ejr{the generated} documents underpinned by the  graph structure and the logical structure are tailored to the specific characteristics of the given ontology and \mr{these embeddings are more effective for the prediction tasks of these ontologies.}

\mr{
\subsection{Classifiers}\label{sec:classifier}
Table \ref{res:classifier} presents the results of four different binary classifiers that use the \owlvec embeddings as input.
Note we used the setting of $D_{s,l}$ + $V_{word}$ because it achieves the best performance on subsumption prediction for FoodOn and GO, and very competitive performance on membership prediction for HeLis in our above evaluation.
We can find that MLP (with single hidden layer) has quite competitive performance as RF, especially on HeLis and FoodOn.
The performance of \owlvec with MLP is also better than all the baselines in Table \ref{res:overall} on each ontology.
SVC can work for HeLis and GO, but it has lower MRR and Hits@1 than RF and MLP.
On the other hand, we find LR, the only linear classifier, has very poor performance on all the three ontologies.
This indicates that the embeddings learned by \owlvec would need some non-linear classifiers attached for good performance in membership prediction and subsumption prediction.
}

\begin{table}[t]
\scriptsize{
\renewcommand{\arraystretch}{1.3}
\centering
\begin{tabular}[t]{p{1.4cm}<{\centering}||p{1cm}<{\centering}|p{1cm}<{\centering}||p{1cm}<{\centering}|p{1cm}<{\centering}||p{1cm}<{\centering}|p{1cm}<{\centering}||p{1cm}<{\centering}|p{1cm}<{\centering}}\hline
&\multicolumn{2}{c||}{RF} & \multicolumn{2}{c||}{MLP}  & \multicolumn{2}{c||}{SVC} & \multicolumn{2}{c}{LR} \\\cline{2-9}
& MRR & Hits@$1$ & MRR & Hits@$1$ & MRR & Hits@$1$ & MRR & Hits@$1$\\ \hline
HeLis &$0.938$ &$0.920$ &$0.935$ &$0.916$ &$0.816$ &$0.752$ &$0.135$ &$0.052$ \\ \hline
FoodOn &$0.213$ &$0.143$ &$0.219$ &$0.120$ &$0.063$ &$0.033$ &$0.012$ &$0.009$ \\ \hline
GO &$0.170$ &$0.076$ &$0.152$ &$0.064$ &$0.138$ &$0.071$ &$0.011$ &$0.008$ \\ \hline
\end{tabular}
\vspace{-0.1cm}
\caption{\footnotesize
\mr{Performance of the classifiers of Random Forest (RF), Multi-Layer Perception (MLP), Support Vector Classifier (SVC) and Logistic Regression (LR), using \owlvec ($D_{s,l}$ + $V_{word}$).}
}\label{res:classifier}
}
\end{table}

\subsection{Interpretation and Visualization}\label{sec:visualization}
\begin{figure}[t]
\begin{center}
\includegraphics[width=0.95\textwidth]{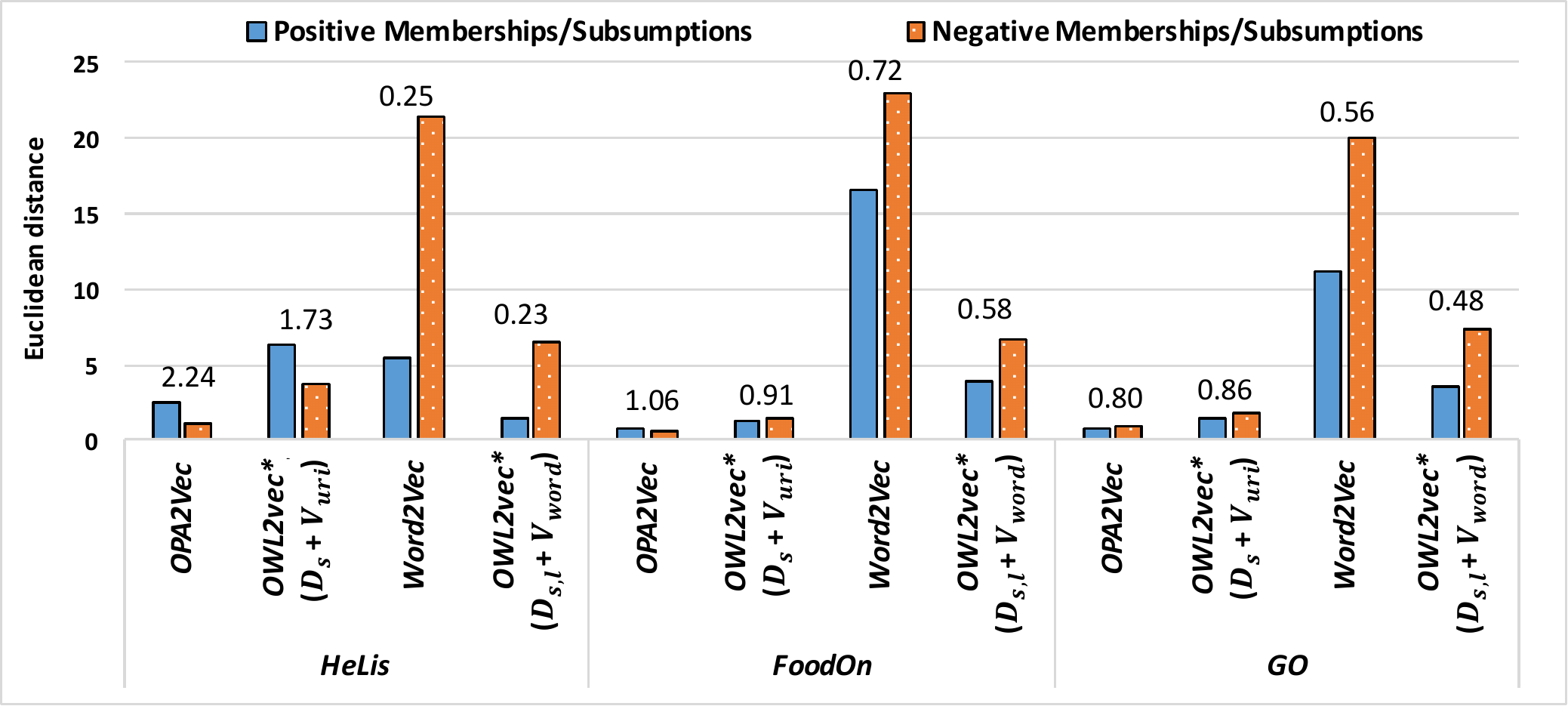}
\vspace{-0.1cm}
\caption{\footnotesize
The average Euclidean distance between the class and its instance (resp. subclass) for the the positive and negative memberships (resp. subsumptions) used in classifier training. The number above every pair of positive and negative bars is their ratio.
}\label{res:distance}
\end{center}
\end{figure}

To show that the learned embeddings (i.e., input features of the classifier for membership and subsumption prediction) are discriminative and effective, we analyze the Euclidean distance between the embeddings of the two entities in a membership or subsumption axiom. 
We calculate the average distance for \mr{the true axioms that are extracted from the ontology, and the false axioms that are constructed by corrupting each true axiom (i.e., the same way as negative sampling in the case studies),
using} the embeddings learned by OPA2Vec, the pre-trained \textit{Word2Vec}, and \owlvec with two settings.
The results are shown in Fig.~\ref{res:distance}.
Note that the difference of the Euclidean distance between the entities in the positive axioms and the entities in the negative axioms is 
\ejr{sufficient}
to indicate the discrimination of the features, but it is not necessary. 
We can find that \textit{Word2Vec} and \owlvec with $D_{s,l}$ + $V_{word}$ (\ie using the structure document, the lexical document and the word embedding)  have quite discriminative average distances for all the three ontologies. 
Namely, the positive axioms lead to much shorter average distance than the negative axioms. 
This is consistent with their final good performance shown above.
Specially, for OPA2Vec and \owlvec with $D_{s}$ + $V_{iri}$ (\ie using the structure document and the IRI embedding) on HeLis, we can find the distance is also discriminative.
However, in \mr{contrast}, the positive axioms has longer average distance than the negative axioms.
This is because the instance usually lies in one end of a sequence where it co-occurs with its class (\ie a walk of WL sub-tree kernel of depth $3$ for \owlvec, or a membership axiom for OPA2Vec), and thus its distance of co-occurrence to its class becomes larger than to a random class.  

\begin{figure}[t]
\begin{center}
\begin{subfigure}[b]{\textwidth}
\includegraphics[width=0.99\textwidth]{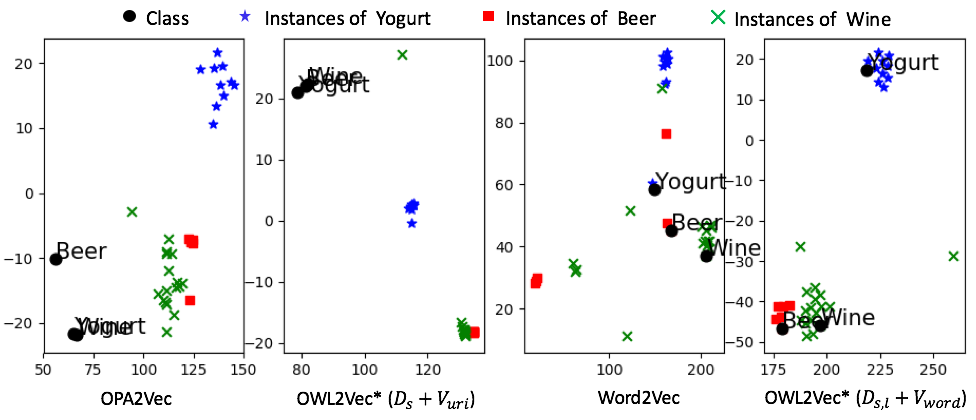}
\caption{
Three classes of HeLis -- ``Yogurt'', ``Beer'' and ``Wine'', and their instances ($10$, $6$ and $17$ respectively).
}\label{res:helis_visualization}
\vspace{0.2cm}
\end{subfigure}
\begin{subfigure}[b]{\textwidth}
\includegraphics[width=0.99\textwidth]{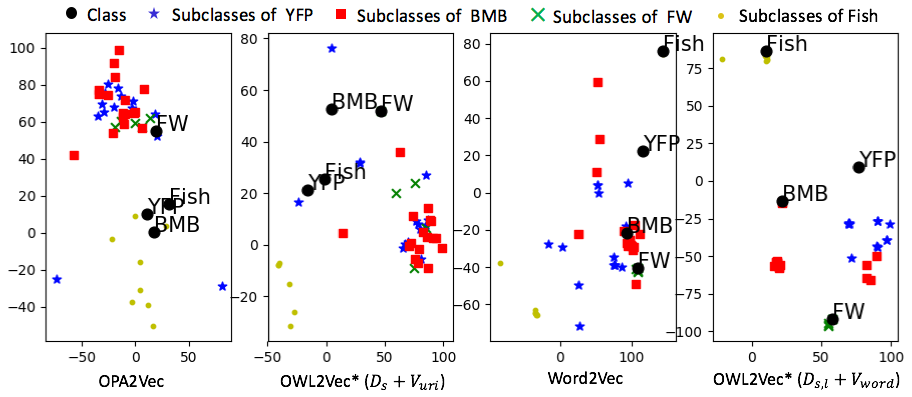}
\caption{
Four classes of FoodOn -- ``Yogurt Food Product'' (YFP), ``Barley Malt Beverage'' (BMB), ``Fruit Wine'' (FW) and ``Fish'', and their subclasses ($14$, $17$, $5$ and $8$ respectively).
}\label{res:foodon_visualization}
\end{subfigure}
\vspace{-0.4cm}
\caption{\footnotesize Embedding visualization via t-SNE.}\label{res:visualization}
\end{center}
\end{figure}

We also visualize the embeddings of some example classes/instances via t-SNE \cite{maaten2008visualizing} \ejr{in order to obtain further insights about the quality of the computed embeddings}.
In Figure~\ref{res:helis_visualization} (for HeLis) we can find two characteristics for the embeddings learned by \owlvec with $D_{s,l}$ and $V_{word}$: \textit{(i)} the instances of each class are clustered into a compact cluster, and \textit{(ii)} these instances are very close to their corresponding class.
Both characteristics are promising: they verify that the embeddings are discriminative and explain why the embeddings enable a very good performance in membership prediction (e.g., Hits@$5$ is as high as $0.978$).
For the embeddings learned by OPA2Vec and \owlvec with $D_{s}$ and $V_{iri}$, they have the first characteristic as well, but the distance of an instance to its class is often longer than its distance to some other class, which is consistent with the average Euclidean distance analyzed above. 
Such embeddings can still benefit membership prediction under the standard supervised learning setting adopted in our evaluation, where some instances of one class are used for training while the other instances of this class, which are close to the training instances in the embedding space, are for testing.
However, the generalization will be dramatically impacted, especially under a zero-shot learning setting where the instances of a new class, which have never appeared in the training samples, are used for testing.

In Figure \ref{res:foodon_visualization} (for FoodOn) we can observe similar characteristics for the embeddings learned by \owlvec with $D_{s,l}$ and $V_{word}$. 
Namely, for each class, its subclasses are mostly quite close to each other (i.e., being clustered into one cluster), and their distances to this class are mostly shorter than their distance to any other class.
However, the two characteristics are not as significant as in HeLis, especially for the class ``Barley Malt Beverage'' and its subclasses, indicating that embedding FoodOn, which has more axioms and entities (cf. Table \ref{res:statistics}), is more challenging.
On the other hand, the two characteristics of \owlvec with $D_{s,l}$ and $V_{word}$ are more significant than those of the other three methods --- \textit{Word2Vec}, OPA2Vec and \owlvec with $D_{s}$ and $V_{iri}$, which verifies its better performance on subsumption prediction. 
For example, in comparison with \textit{Word2Vec} which has the second best performance, \owlvec with $D_{s,l}$ and $V_{word}$ closes the distance between ``Fish'' and its subclasses, and makes the subclasses of ``Yogurt Food Product'' closer to each other.

\vspace{-0.3cm}

\section{Discussion and Outlook}\label{sec:conclusion}
In this paper we have presented \owlvec, a robust semantic embedding framework for \owl ontologies.
\owlvec extracts documents from the ontology that capture its graph structure, axioms of logical constructors, as well as its lexical information,
and then learns a \mr{word embedding} model for both entity embeddings and word embeddings.
We applied \owlvec to class membership prediction and class subsumption prediction with three real world ontologies, namely HeLis, FoodOn and GO, and we empirically analysed different semantics and techniques such as entailment reasoning and ontology to RDF graph transformation.
The evaluation demonstrates that on these tasks \owlvec can significantly outperform state-of-the-art methods.

\medskip
\noindent
\textbf{Ontology Text Understanding. }
Our experiments suggest that lexical information plays a very important role in both class membership prediction and class subsumption prediction.
In real world ontologies such as HeLis, FoodOn and GO, entity names often reflect, in natural language, their relationships to surrounding entities; in HeLis, for example, the instance \textit{vc:FOOD-700637} (\textit{Soy Milk}) is an 
instance of the class \textit{vc:SoyProducts}.
In addition, ontologies often contain a large number of entity annotations ranging from short phrases to long textual descriptions.
In FoodOn, for example, $169,630$ out of $241,581$ axioms are annotations.
However, patterns within the textual information in the ontologies, which is underpinned by the graph and logical structure, are quite different from normal natural language text (cf. Section \ref{sec:pt}).
To further improve ontology embedding in the future, we need to develop new \mr{word embedding} architectures and training methods that are tailored to the kinds of textual information typically present in state-of-the-art ontologies.

\medskip
\noindent
\textbf{Ontology Completion via Prediction. }
In this study \owlvec has been used to complete an ontology by discovering plausible axioms.
We adopted a typical supervised learning setting to model a common scenario in ontology completion, where satisfactory results have been achieved; in class membership prediction, the classes of $93.2\%$ of the test instances can be recalled.
In some real world cases,
however, there is often a bias between the axioms for training and the  axioms for prediction.
For example, consider the case of
membership prediction for a new class defined on the fly without any known instances (i.e., zero-shot learning scenario discussed Section \ref{sec:visualization}).
This leads to sample shortage in training and becomes much more challenging --- the above metric drops to $65.6\%$ for \owlvec and less than $10\%$ for other KG embedding and ontology embedding methods in Table~\ref{res:overall}.
In future work we plan to develop more robust ontology embeddings with higher generalization for dealing with such cases.
\mr{Meanwhile, we will consider using \owlvec embeddings and machine learning to address other ontology completion challenges. One study we are working on is predicting the cross-ontology class mapping for ontology integration and curation \cite{chen2020augmenting,horrocks2020tool};
while the other potentially meaningful study is approximating the result of those logical deductive inference tasks over expressive ontologies, which have exponential or even higher time complexity by the traditional logical reasoners.}

\medskip
\noindent
\mr{
\textbf{Industrial Applications. }
Besides the evaluated membership and subsumption prediction, \owlvec can be applied to assist a wide range of ontology design and quality assurance (QA) problems \cite{horrocks2020tool}.
A typical QA task is ontology alignment as presented in our ongoing work \cite{chen2020augmenting}, where we use \owlvec to embed the classes of two to-be-aligned ontologies as their features for mapping prediction.
Note through the ontology mappings, we can further use the cross-ontology information to augment subsumption and membership prediction; for example, the missing subsumption relationship between \textit{obo:FOODON\_03305289 (Soybean Milk)} and \textit{obo:FOODON\_00002266 (Soybean Food Product)} in FoodOn (where \textit{obo:FOODON\_03305289} is only categorized as \textit{obo:FOODON\_00003202 (Beverage)}) can be discovered by mapping them to their HeLiS counterparts \textit{vc:SoyMilk} and \textit{vc:SoyProducts} whose subsumption relationship is defined. 
Both \owlvec and \cite{chen2020augmenting} are in cooperation with Samsung Research UK, aiming at building and curating a high quality food ontology which is beneficial to artificial intelligence and information systems in domains such as personal health and agriculture.
The entity clustering by \owlvec can also contribute to ontology design by e.g., discovering potential classes that have not been defined, as well as ontology QA by e.g., entity resolution.
On the other hand, \owlvec, as an ontology tailored word embedding model, could replace the original word embedding models and would work better for some domain specific tasks such as biomedical text analysis \cite{hao2020enhancing}.
This is also a promising direction worth studying.
}

\vspace{-0.3cm}
\section*{Acknowledgments}
\vspace{-0.3cm}
This work was supported by the SIRIUS Centre for Scalable Data Access (Research Council of Norway, project 237889), Samsung Research UK, Siemens AG, and the EPSRC projects AnaLOG (EP/P025943/1), OASIS (EP/S032347/1), UK FIRES (EP/S019111/1) and the AIDA project (Alan Turing Institute).
\vspace{-0.5cm}


%
%
\bibliographystyle{splncs04}
\bibliography{reference}

\end{document}